\documentclass[sigconf, nonacm]{acmart}
\AtBeginDocument{%
  }

\usepackage{booktabs}
\usepackage{multirow}
\usepackage{makecell}
\usepackage{graphicx}
\usepackage{caption}
\usepackage{stfloats}
\usepackage{subcaption}
\usepackage{float}

\begin{document}

\title{STaT: Resolving Shape Distortion in Non-Stationary Time Series via Tri-Modal Synergy}

\author{Hui Cheng}
\email{huicheng@mail.hfut.edu.cn}
\affiliation{%
  \institution{Hefei University of Technology}
  \city{Hefei}
  \country{China}
}
\author{Jinsheng Guo}
\email{guojinsheng@mail.hfut.edu.cn}
\affiliation{%
  \institution{Hefei University of Technology}
  \city{Hefei}
  \country{China}
}
\author{Zhenhao Weng}
\email{zhenhaoweng@mail.hfut.edu.cn}
\affiliation{%
  \institution{Hefei University of Technology}
  \city{Hefei}
  \country{China}
}
\author{Yan Qiao}
\email{qiaoyan@hfut.edu.cn}
\authornote{Corresponding authors}
\affiliation{%
  \institution{Hefei University of Technology}
  \city{Hefei}
  \country{China}
}

\author{Meng Li}
\email{mengli@hfut.edu.cn}
\authornotemark[1]
\affiliation{%
  \institution{Hefei University of Technology}
  \city{Hefei}
  \country{China}
}

\begin{abstract}

Recent research in time series forecasting frequently investigates the integration of textual and visual modalities with numerical models to better navigate non-stationary environments. Despite delivering solid numerical results, existing multi-modal approaches usually encounter a dilemma: prioritizing the minimization of average errors can result in excessively smooth forecasts that overlook essential fluctuations. To resolve this limitation, we introduce STaT, an innovative multimodal architecture for Symbolic-Temporal-Textual Alignment, which seamlessly unites three synergistic modalities. Specifically, the symbolic modality converts continuous time series into discrete tokens, facilitating the accurate identification of structural patterns and turning points; the temporal modality extracts inherent sequential dependencies; and the textual modality leverages domain semantics to steer the macroscopic forecasting trends. Comprehensive evaluations on eight real-world benchmarks indicate that STaT delivers exceptional performance, enhancing conventional magnitude indicators by up to 8.9\% while simultaneously decreasing shape distortion by up to 8.5\%.

\end{abstract}



\maketitle

\section{Introduction}
\label{sec:intro}

Time series forecasting is essential for decision-making in complex dynamical systems, with applications across finance \cite{li2018stock, huang2020deep}, weather \cite{nguyen2023climax, gao2022earthformer}, traffic \cite{li2017diffusion, yu2017spatio}, and energy \cite{wan2013probabilistic, runge2021review}. Deep learning paradigms, particularly Transformer-based architectures \cite{zhou2021informer, wu2021autoformer, zhou2022fedformer, nie2022time}, have advanced time series analysis by capturing long-range dependencies. However, single-modality numerical models still struggle to handle non-stationary distribution shifts in long-term forecasting, often resulting in fragile generalization \cite{kim2021reversible, liu2022non, liu2023adaptive}.

\begin{figure}[htbp]
\centering
\includegraphics[width=\columnwidth]{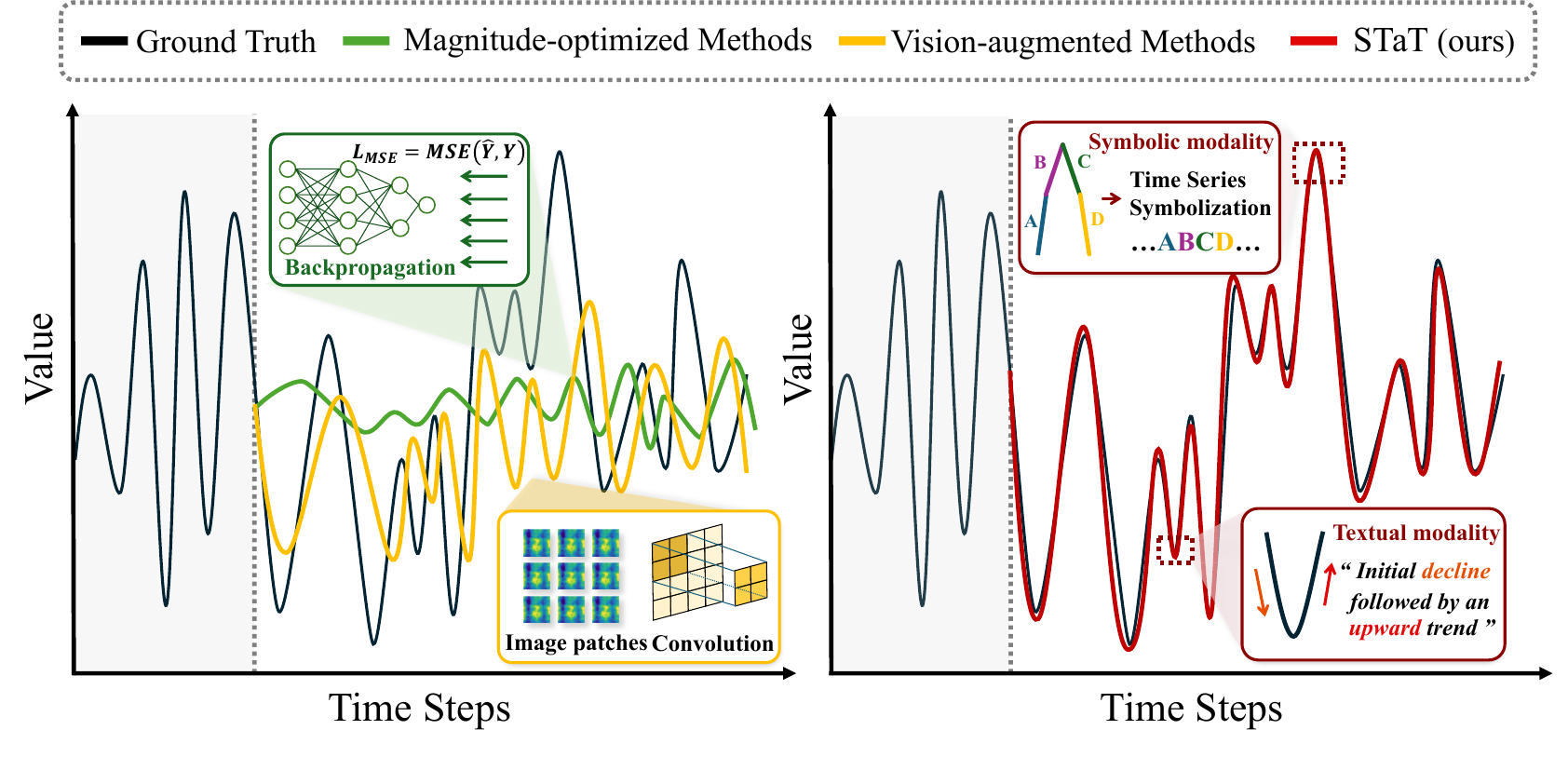}
\caption{Comparison between existing forecasting paradigms and our approach (STaT). \textbf{(Left)} Current paradigms often suffer from over-smoothed predictions. \textbf{(Right)} Our STaT integrates temporal, textual, and symbolic modalities to minimize magnitude errors while ensuring precise shape alignment.}
\label{fig:motivation}
\end{figure}

Recent research explores multi-modal-based forecasting by introducing additional modalities beyond numerical data to adapt to non-stationary environments. Initial efforts augment numerical data with large language models (LLMs) \cite{jin2024time, chang2023llm4ts, xue2023promptcast, gruver2023large} to leverage semantic contexts. However, discrete text alone is too abstract to capture the variation patterns of temporal dynamics. Emerging vision–language models and vision-based techniques have been introduced to transform 1D time series into 2D images, such as heatmaps or line plots, exploiting visual features to complement textual semantics \cite{wang2015encoding, zhong2025time}.

Despite progress in standard magnitude metrics (e.g., mean squared error (MSE) and mean absolute error (MAE)) on non-stationary datasets, current multi-modal-based forecasting paradigms are caught in a trade-off: in pursuit of reducing the average magnitude error, they often forgo capturing volatility in certain instances, resorting instead to smooth predictions to avoid the risk of mispredicting peaks and troughs — since any error at peaks or troughs would cause a sharp increase in magnitude errors \cite{cuturi2017soft, le2019shape} (as shown in Figure~\ref{fig:motivation}, left, green line).

Nevertheless, in numerous real-world applications, the precise identification of peaks and troughs is paramount, even outweighing the accuracy of the mean prediction \cite{laptev2017time, ding2019modeling}. For instance, precisely capturing the turning points of financial asset prices can help avert systemic financial crises; accurately tracking dynamic congestion trends enables proactive traffic diversion; and detecting extreme peak amplitudes of weather anomalies allows for the early deployment of countermeasures against extreme weather events. These real‑world demands underscore the need to prioritize shape alignment alongside magnitude accuracy.

Upon deeper investigation, besides the pursuit of minimizing the magnitude errors, the state-of-the-art visual representation for time series also leads to overly smooth prediction curves. Standard 2D visual operations, such as image patching and convolutions, inherently act as spatial low-pass filters \cite{wu2022timesnet, wang2015encoding} that discard high-frequency mutations (Figure~\ref{fig:motivation}, left, yellow line). Consequently, while these models may successfully achieve even lower magnitude errors, they often suffer from shape distortion.

Symbolic representation is an effective alternative form of sequence representation \cite{lin2003symbolic}. Distinct from the image patching and convolutions used in visual representations, symbolic time series analysis (STSA) discretizes continuous signals into discrete symbols \cite{chen2024fabba}. With the distinct advantage of preserving local shapes, STSA tightly binds geometric structures to their exact temporal positions, effectively resolving shifts in turning points. Unlike visual representations, which have received considerable attention in the rapidly growing field of computer vision, symbolic representation of time series has been relatively underexplored.

In this paper, we introduce, for the first time, a symbolic modality into time series forecasting, complementing the textual and temporal modalities. Specifically, the symbolic modality captures volatility and turning points by encoding numerical series into discrete, interpretable symbols that make abrupt changes structurally explicit; the textual modality captures global trends by incorporating contextual knowledge from external text, providing the overarching direction of the series; and the temporal modality captures temporal correlations, modeling the sequential dependencies and time-varying patterns inherent in the data. To fully empower the three modalities, we design a volatility-aware temperature (VAT) routing mechanism that adaptively adjusts their weights across various non-stationary environments. Through the synergy of the three modalities, our temporal-text-symbolic multimodal architecture, STaT, achieves accurate shape alignment while maintaining a low magnitude error (Figure~\ref{fig:motivation}, right, red line).

In summary, this paper makes the following contributions to time series forecasting.
\begin{itemize}
\item We propose the first multimodal framework that integrates symbolic, textual, and temporal modalities. This new framework effectively achieves minimal magnitude errors while simultaneously ensuring precise shape alignment to capture exact turning points, preserve extreme peak amplitudes, and track consistent trends.

\item We design a volatility-aware temperature (VAT) routing mechanism that adaptively adjusts the fusion weights of the three modalities based on the current volatility of the time series. This mechanism maximizes their synergistic complementarity across diverse non-stationary environments, contributing to robust performance across various volatility scenarios.

\item We conduct extensive experiments on eight real-world datasets. Empirical results demonstrate that while existing baselines struggle to balance magnitude accuracy and shape alignment, STaT achieves breakthrough performance on both fronts. It improves standard magnitude metrics by up to 8.9\% while simultaneously reducing shape distortion by up to 8.5\%.

\end{itemize}

\section{Related Work}
This section reviews three lines of time series forecasting research closely related to this work: multimodal-based forecasting, symbolic-based forecasting, and shape-aligned forecasting.

\subsection{Multi-Modal Forecasting}
Multi-modal architectures drive new paradigms in time series forecasting. GPT4TS \cite{zhou2023one} and TimeLLM \cite{jin2024time} project temporal patches \cite{nie2022time} into textual representations. TimeCMA \cite{liu2025timecma} aligns temporal features with textual prompts. Recent studies expand these integrations to include visual modality. Time-VLM \cite{zhong2025time} utilizes vision-language models to enrich temporal representations. T3Time \cite{chowdhury2026t3time} integrates temporal, spectral, and prompt-based features. However, these paradigms commonly rely on magnitude error as the loss function or on visual operations, which may introduce averaging biases and low-pass filtering effects, often resulting in over-smoothed predictions that overlook critical volatility.

\subsection{Symbolic Representation}
Symbolic representation establishes a crucial bridge between continuous numerical sequences and discrete semantic spaces. Early frameworks, including SAX \cite{lin2007experiencing} and 1d-SAX \cite{malinowski20131d}, map temporal data into finite characters via fixed-window aggregation and linear trend extraction. Subsequent adaptive methods, such as ABBA \cite{elsworth2020abba} and fABBA \cite{chen2023efficient}, dynamically model time series patterns through joint amplitude and period variations using polygonal chain approximations. A recent study, LLM-ABBA \cite{chen2024llm}, aligns these discrete symbols directly with the native vocabularies of pre-trained language models. However, despite their diverse symbolization strategies, relying solely on the symbolic modality makes it difficult to perceive overarching trends and long-range dependencies.

\subsection{Shape-Aligned Forecasting}
Shape-aligned forecasting aims to preserve temporal dynamics by adopting shape-focused losses such as Soft-DTW \cite{cuturi2017soft} and DILATE \cite{le2019shape}, instead of simple magnitude error minimization. Dong et al. \cite{dong2025teaching} proposed a multimodal contrastive framework that optimizes InfoNCE-based alignment loss to reconstruct structural dependencies from visual and textual perspectives. Wang et al. \cite{wang2025time} introduced a transformation-augmented objective, Time-o1, which projects label sequences into decorrelated components via singular value decomposition (SVD). Kudrat et al. \cite{kudrat2025patch} proposed a patch-wise structural (PS) loss that integrates localized correlation, variance, and mean metrics to produce more accurate shape predictions. However, without explicit discrete structural features, these paradigms—operating in continuous spaces—remain susceptible to filtering biases, making it challenging to strike a balance between shape constraints and magnitude accuracy.

In contrast to the above methods, the proposed STaT integrates the symbolic, textual, and temporal modalities, allowing them to complement one another. We further design a VAT routing mechanism coupled with an adaptive dual fidelity (ADF) loss tailored to these modalities, thereby enabling the predictions to achieve jointly optimal magnitude accuracy and shape alignment.

\begin{figure*}[htbp]
\centering
\IfFileExists{framework.pdf}{\includegraphics[width=\textwidth]{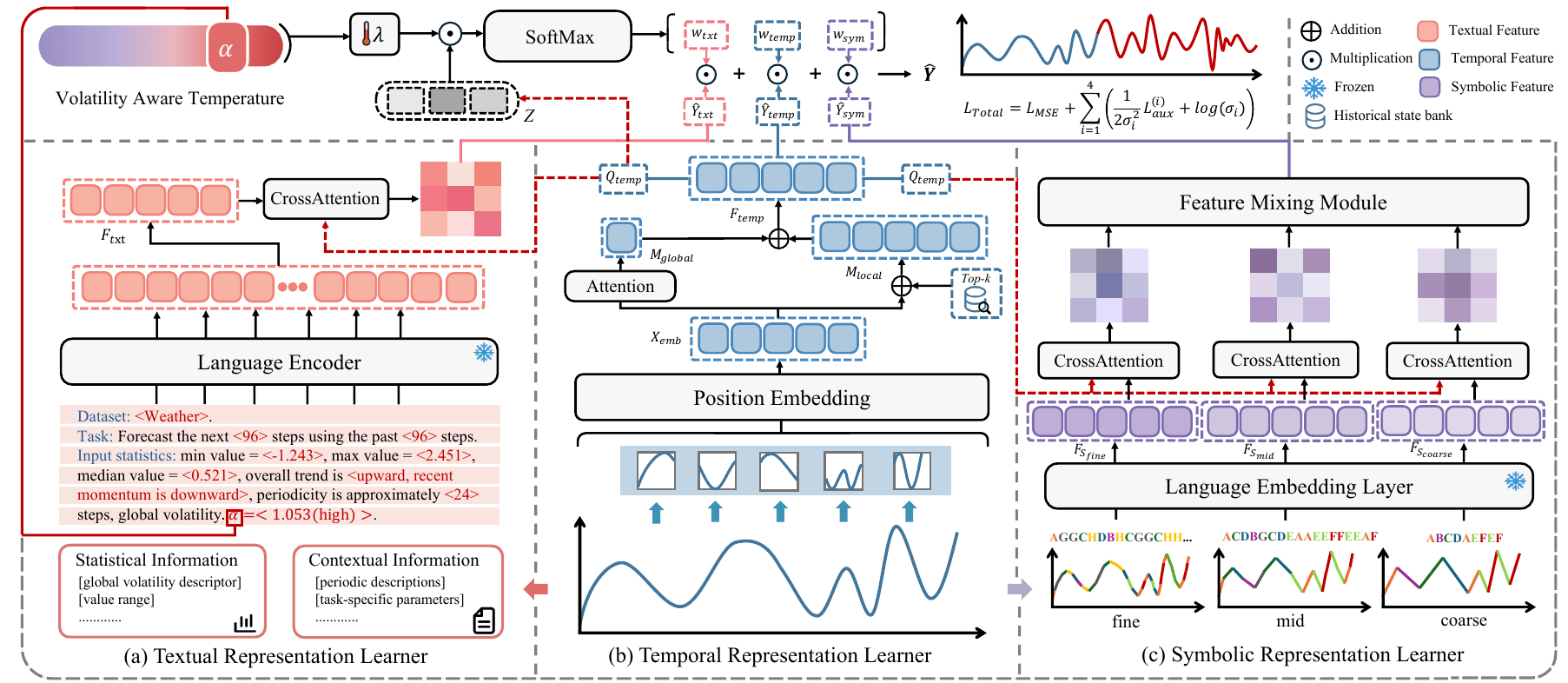}}{\rule{\textwidth}{6cm}}
\caption{Overview of the STaT framework.}
\label{fig:framework}
\end{figure*}

\section{Methodology}

\textbf{Problem Formulation.} We consider a multivariate time series forecasting task. Let the historical observations be denoted as $\mathbf{X} \in \mathbb{R}^{B \times L \times C}$, where $B$ represents the batch size, $L$ is the look-back window length, and $C$ denotes the number of variates. The goal is to predict the future sequence $\hat{\mathbf{Y}} \in \mathbb{R}^{B \times T \times C}$, where $T$ is the forecasting horizon. We define our multi-modal forecasting model as a mapping function $\mathcal{F}_\Theta: \mathbb{R}^{B \times L \times C} \rightarrow \mathbb{R}^{B \times T \times C}$ parameterized by $\Theta$. Given a historical sequence $\mathbf{X} \in \mathbb{R}^{B \times L \times C}$, our goal is to accurately forecast its future values $\mathbf{Y} \in \mathbb{R}^{B \times T \times C}$ over the next $T$ time steps. The forecasting model $\mathcal{F}_\Theta$ is trained to tightly align the predicted sequence $\hat{\mathbf{Y}}$ with the ground truth $\mathbf{Y}$ by concurrently minimizing the magnitude error and preserving shape fidelity.

\textbf{Method Overview.} To achieve this goal, we propose STaT. As illustrated in Figure~\ref{fig:framework}, our framework integrates three components to utilize their strong complementarity: a \textit{1) temporal representation learner} to establish base predictions and queries, a \textit{2) textual representation learner} providing macroscopic semantic priors, and a \textit{3) symbolic representation learner} extracting multi-scale discrete geometric structures. These multi-modal representations are dynamically harmonized via a \textit{VAT routing mechanism} and optimized end-to-end using the \textit{ADF Loss}.

\subsection{Temporal Representation Learner}
In the temporal representation learner, the input data is first tokenized into latent vectors via \textit{patch projection}, then fused with retrieved historical states via \textit{historical auto-correlation modeling}, and finally projected to future horizons and cross-modal dimensions via \textit{prediction and query generation} to serve as queries for the other modalities.

\textbf{Patch Projection:} To capture local temporal patterns, the input sequence $\mathbf{X}$ is first partitioned into overlapping patches $\mathbf{X}_{p} \in \mathbb{R}^{B \times N \times (P \cdot C)}$, where $P$ denotes the patch length and $N$ is the total number of patches. These patches are then linearly projected into a high-dimensional latent space $\mathbb{R}^{D_{\mathrm{model}}}$ and injected with positional embeddings to preserve temporal ordering, resulting in the embedded sequence $\mathbf{X}_{\mathrm{emb}} \in \mathbb{R}^{B \times N \times D_{\mathrm{model}}}$.

\textbf{Historical Auto-Correlation Modeling:} To learn both local and global autocorrelation features of the time series, we construct a unified reference pool that captures complex temporal dynamics by maintaining a historical state bank $\mathbf{B}_{\mathrm{mem}} \in \mathbb{R}^{M \times D_{\mathrm{model}}}$, where $M$ denotes the maximum bank capacity. 
During each forward pass, the current patch embeddings are first temporally averaged and then enqueued into $\mathbf{B}_{\mathrm{mem}}$, while the oldest representations are dynamically dequeued to keep the bank up‑to‑date. The extraction of autocorrelations proceeds at two levels, local and global:

\quad \textit{\textbf{Local Auto-Correlation:}} We retrieve the top-$k$ similar historical patches from the state bank based on their cosine similarity with the current embeddings $\mathbf{X}_{\mathrm{emb}}$. These retrieved patches are processed through a two-layer MLP and combined with the original $\mathbf{X}_{\mathrm{emb}}$ via a residual connection to extract local auto-correlation features:
\begin{equation}
\mathbf{M}_{\mathrm{local}} = \mathbf{X}_{\mathrm{emb}} + \text{MLP}(\text{Top-}k(\mathbf{B}_{\mathrm{mem}})).
\label{eq:m_local}
\end{equation}

\quad \textit{\textbf{Global Auto-Correlation:}} To capture long-range dependencies, we apply multi-head self-attention over linear projections of the current patch embeddings to yield contextualized representations. The global auto-correlation $\mathbf{M}_{\mathrm{global}}$ is then obtained by temporally averaging these contextualized attention outputs across the patch dimension $N$:
\begin{equation}
\mathbf{M}_{\mathrm{global}} = \frac{1}{N} \sum_{i=1}^{N} \text{Attn}(\mathbf{X}_{\mathrm{emb}})_i.
\label{eq:m_global}
\end{equation}
The two representations are fused via broadcasting addition to capture high-level temporal patterns $\mathbf{F}_{\mathrm{temp}} = \mathbf{M}_{\mathrm{local}} + \mathbf{M}_{\mathrm{global}}$.

\textbf{Prediction and Query Generation:} The temporal representation $\mathbf{F}_{\mathrm{temp}} \in \mathbb{R}^{B \times N \times D_{\mathrm{model}}}$ serves a dual purpose. It is processed by a linear head to generate the independent temporal representation $\hat{\mathbf{Y}}_{\mathrm{temp}} \in \mathbb{R}^{B \times T \times C}$, and is simultaneously transformed via a normalized linear projection into a cross-modal query axis $\mathbf{Q}_{\mathrm{temp}} \in \mathbb{R}^{B \times C \times D_{\mathrm{model}}}$ utilized for textual and symbolic space retrieval.
 
\subsection{Textual Representation Learner}
\label{sec:textual_learner}
In the textual representation learner, the numerical input is first translated into descriptive prompts via \textit{dynamic prompting}, then mapped into a high-dimensional semantic space via \textit{textual feature extraction}, and finally fused with temporal queries to generate text-augmented forecasts via \textit{textual prediction}.

\textbf{Dynamic Prompting:} As shown in Table~\ref{tab:prompt_example}, we transform the raw data $\mathbf{X}$ into a textual prompt $\mathbf{X}_{\text{text}}$, which consists of a \textit{task specification} section and a \textit{dynamic statistics} section. The \textit{task specification} contextualizes the forecasting objective by specifying the dataset and the prediction horizon. The \textit{dynamic statistics} section provides precise numerical grounding through statistical properties such as trend, momentum, and global volatility. 

To enable the model to adapt to different volatility regimes, we extract a \textit{volatility descriptor} $\alpha$ for the current input window, defined as:
\begin{equation}
\label{eqn:descriptor}
\alpha = \frac{1}{C}\sum_{c=1}^{C}\text{std}(\mathbf{x}_c),
\end{equation}
where $\text{std}(\mathbf{x}_c)$ denotes the standard deviation of the $c$-th individual variate ($c \in \{1, 2, \dots, C\}$) within the normalized input window.

\begin{table}[ht]
  \centering
  \caption{Structural template of the dynamic textual prompt $\mathbf{X}_{\mathrm{text}}$.}
  \label{tab:prompt_example}
  \small
  \renewcommand{\arraystretch}{1.3}
  \begin{tabular}{@{}p{0.95\columnwidth}@{}}
    \toprule
    \textbf{[Task Specification]} \\
    \textbf{Dataset:} \textit{$<$domain-specific dataset description$>$} \\
    \textbf{Task:} Forecast the next \textit{$<$$T$$>$} steps using the past \textit{$<$$L$$>$} steps. \\
    \midrule
    \textbf{[Dynamic Statistics]} \\
    \textbf{Input statistics:} min value = \textit{$<$min$>$}, max value = \textit{$<$max$>$}, median value = \textit{$<$med$>$}, overall trend is \textit{$<$upward$|$downward$>$}, recent momentum is \textit{$<$upward$|$downward$>$}, periodicity is approximately \textit{$<$$P$$>$} steps, \\
    \textbf{Volatility descriptor $\alpha$} = \textit{$<$value$>$} (\textit{$<$high$|$moderate$|$low$>$}). \\
    \bottomrule
  \end{tabular}
\end{table}

\textbf{Textual Feature Extraction:} The constructed text sequence $\mathbf{X}_{\mathrm{text}}$ is fed into a pre-trained language encoder to produce high-dimensional textual embeddings, which are then aligned to the model's primary dimension:

\begin{equation}
\mathbf{F}_{\mathrm{txt}} = \text{Linear}_{\mathrm{align}}(\text{Embed}_{\mathrm{Language}}(\mathbf{X}_{\mathrm{text}}))
\label{eq:f_txt},
\end{equation}
where $\text{Embed}_{\mathrm{Language}}(\cdot)$ denotes the frozen language embedding layer of the pre-trained language encoder, and $\mathbf{F}_{\mathrm{txt}} \in \mathbb{R}^{B \times L_{\mathrm{txt}} \times D_{\mathrm{model}}}$ forms the macroscopic semantic knowledge pool, with $L_{\mathrm{txt}}$ representing the textual token length.

\textbf{Textual Prediction:} The temporal query $\mathbf{Q}_{\mathrm{temp}}$ attends to the semantic pool via a cross-attention mechanism to generate the text-augmented prediction:
\begin{equation}
\mathbf{O}_{\mathrm{txt}} = \text{CrossAttention}(\mathbf{Q}=\mathbf{Q}_{\mathrm{temp}}, \mathbf{K}=\mathbf{F}_{\mathrm{txt}}, \mathbf{V}=\mathbf{F}_{\mathrm{txt}}),
\label{eq:o_txt}
\end{equation}
\begin{equation}
\hat{\mathbf{Y}}_{\mathrm{txt}} = \text{GELU}(\text{LayerNorm}(\text{Linear}_{\mathrm{txt\_head}}(\mathbf{O}_{\mathrm{txt}}))),
\label{eq:y_txt}
\end{equation}
where $\mathbf{O}_{\mathrm{txt}} \in \mathbb{R}^{B \times C \times D_{\mathrm{model}}}$ is the retrieved semantic feature, and $\hat{\mathbf{Y}}_{\mathrm{txt}} \in \mathbb{R}^{B \times T \times C}$ is the resulting textual representation.

\subsection{Symbolic Representation Learner}
In the symbolic representation learner, the continuous sequence is first discretized into multi-scale discrete symbols via \textit{time series symbolization}, then projected into a continuous semantic space via \textit{symbolic feature extraction}, and finally queried across scales to formulate symbolic representation via \textit{symbolic prediction}.

\textbf{Time Series Symbolization:} We first apply window-level Z-score normalization to the continuous sequence $\mathbf{X}$ to ensure scale consistency, and yield the normalized sequence $\tilde{\mathbf{X}}$. Then we employ the fABBA algorithm \cite{chen2024fabba} to discretize it into piecewise linear approximations. The algorithm first segments the original time series into a sequence of 2D tuples $(len_i, inc_i)$ representing the length and increment of each piece, ensuring that the reconstruction error remains strictly bounded by a predefined tolerance $tol$. Next, a distance-based clustering mechanism groups these 2D tuples to construct a discrete vocabulary $\Sigma$. Each cluster center $c_k$ encapsulates a unique local geometric structure and is assigned a distinct alphabetical identifier (e.g., 'A', 'B', 'C', 'D'). The mapping function can be formally defined as:
\begin{equation}
v_i = \arg\min_{c_k \in \mathcal{C}} \| (len_i, inc_i) - c_k \|_2,\quad v_i \in \Sigma.
\label{eq:symbol_mapping}
\end{equation}
By replacing each linear segment with its corresponding identifier $v_i$, $\tilde{\mathbf{X}}$ is completely transformed into a discrete symbolic string $\mathbf{S} = [v_1, v_2, \dots, v_{L_{\mathrm{sym}}}]$, where $L_{\mathrm{sym}}$ denotes the length of the symbolic sequence. We denote this entire STSA pipeline as the function $\text{fABBA}(\tilde{\mathbf{X}}, tol)$, where $tol$ is the error tolerance indicating the maximum permissible reconstruction deviation. To systematically capture these discrete symbols at multiple granularities, we extract three distinct symbolic sequences by varying the error tolerance:
\begin{equation}
\mathbf{S}_s = \text{fABBA}(\tilde{\mathbf{X}}, tol_s)
\label{eq:multiscale_symbol}
\end{equation}

where $s \in \{\mathrm{fine, mid, coarse}\}$, and $\mathbf{S}_s \in \Sigma^{L_{\mathrm{sym}}}$ represents the discrete symbolic sequence for scale $s$. $tol_{\mathrm{fine}}$, $tol_{\mathrm{mid}}$, and $tol_{\mathrm{coarse}}$ are configured to capture exact turning points, short-term local fluctuations, and broader amplitude shifts, respectively.

\textbf{Symbolic Feature Extraction:} To treat the alphabetical identifiers as textual tokens, we map the discrete symbolic strings into a continuous high-dimensional vector pool utilizing the frozen language embedding layer of the pre-trained language encoder and the subsequent alignment projection used in \textit{textual feature extraction}:
\begin{equation}
\mathbf{F}_s = \text{Linear}_{\mathrm{align}}(\text{Embed}_{\mathrm{Language}}(\mathbf{S}_s)),
\label{eq:f_sym}
\end{equation}
where $\mathbf{F}_s \in \mathbb{R}^{B \times L_{\mathrm{sym}} \times D_{\mathrm{model}}}$ is the symbolic feature. This projection explicitly aligns the discrete symbols with the textual features in the same representation space, ensuring that the temporal query can seamlessly attend to both global semantic contexts and local geometric structures.

\textbf{Symbolic Prediction:} With the continuous symbolic features $\mathbf{F}_s$ established, the temporal query $\mathbf{Q}_{\mathrm{temp}}$ conducts independent cross-modal retrieval across the three scales to generate the corresponding symbolic representation:
\begin{equation}
\mathbf{O}_s = \text{CrossAttention}(\mathbf{Q}=\mathbf{Q}_{\mathrm{temp}}, \mathbf{K}=\mathbf{F}_s, \mathbf{V}=\mathbf{F}_s),
\label{eq:o_sym}
\end{equation}
\begin{equation}
\hat{\mathbf{Y}}_s = \text{GELU}(\text{LayerNorm}(\text{Linear}_{\mathrm{sym\_head}}(\mathbf{O}_s))),
\label{eq:y_sym_scale}
\end{equation}
where $\mathbf{O}_s \in \mathbb{R}^{B \times C \times D_{\mathrm{model}}}$ is the retrieved structural feature, and $\hat{\mathbf{Y}}_s \in \mathbb{R}^{B \times T \times C}$ is the specific representation for scale $s$.

To adaptively integrate these multi-scale presentations, we use a feature mixing module to dynamically fuse them into a unified symbolic representation $\hat{\mathbf{Y}}_{\mathrm{sym}} = c_{\mathrm{fine}}\hat{\mathbf{Y}}_{\mathrm{fine}} + c_{\mathrm{mid}}\hat{\mathbf{Y}}_{\mathrm{mid}} + c_{\mathrm{coarse}}\hat{\mathbf{Y}}_{\mathrm{coarse}}$ utilizing scale-specific importance weights $[c_{\mathrm{fine}}, c_{\mathrm{mid}}, c_{\mathrm{coarse}}] = \text{Softmax}(\boldsymbol{\omega}_{\mathrm{fmm}})$, where $\boldsymbol{\omega}_{\mathrm{fmm}} \in \mathbb{R}^3$ represents learnable parameters optimized during training to adaptively balance the three structural granularities.

\subsection{Multi-Modal Representation Fusion}
The multi-modal representation fusion module dynamically harmonizes the presentations produced by the temporal, textual, and symbolic learners through our \textit{VAT routing mechanism}, thereby accommodating various volatility scenarios.

In the \textit{VAT routing mechanism}, we first compute the basic step-wise weights  $\mathbf{Z} \in \mathbb{R}^{B \times T \times C \times 3}$ for the three modality experts, by applying a linear projection layer to the pure temporal query $\mathbf{Q}_{\mathrm{temp}}$. To accommodate various volatility scenarios, we modify these weights using the volatility descriptor $\alpha$ derived by Eqn. (\ref{eqn:descriptor}). This descriptor generates an inverse temperature $\lambda$ for the VAT routing:
\begin{equation}
\lambda = \eta \cdot \text{Sigmoid}(\alpha),
\label{eq:VAT_lambda}
\end{equation}
where $\text{Sigmoid}(\cdot)$ is the sigmoid function, and $\eta$ is a hyperparameter that controls the scale of $\lambda$. Here, $\lambda$ directly controls the routing sharpness: low volatility produces a small $\lambda$ for an evenly averaged ensemble, whereas high volatility yields a large $\lambda$ to amplify the dominant modality. The experimental analysis of the VAT mechanism is provided in Section~\ref{sec:model_analysis} and Figure~\ref{fig:mechanism}.

The final forecasting output $\hat{\mathbf{Y}}$ is formulated by aggregating the modality-specific predictions utilizing the VAT-adjusted dynamic fusion weights:
\begin{equation}
[w_{\mathrm{temp}}, w_{\mathrm{txt}}, w_{\mathrm{sym}}] = \text{Softmax}_{\mathrm{dim}=-1}\Big( \lambda \cdot (\mathbf{Z} + \mathbf{b}) \Big),
\label{eq:fusion_weights}
\end{equation}
\begin{equation}
\hat{\mathbf{Y}} = w_{\mathrm{temp}} \odot \hat{\mathbf{Y}}_{\mathrm{temp}} + w_{\mathrm{txt}} \odot \hat{\mathbf{Y}}_{\mathrm{txt}} + w_{\mathrm{sym}} \odot \hat{\mathbf{Y}}_{\mathrm{sym}},
\label{eq:final_output}
\end{equation}
where $\mathbf{b} \in \mathbb{R}^3$ is a constant initialization bias applied to temporarily favor the temporal backbone and stabilize the routing distribution during the initial training phase.

\subsection{Optimization}
Since the three modalities—temporal, textual, and symbolic—capture complementary aspects of the time series, their fusion provides a unified interface for jointly optimizing both magnitude error and shape alignment. However, standard loss functions, such as MSE (e.g., MSE) or shape‑focused losses (e.g., Soft‑DTW), often struggle to balance magnitude accuracy and shape fidelity. 

To address this gap, we design a new training loss function, termed the ADF loss, which is defined as:
\begin{equation}
\mathcal{L}_{\mathrm{Total}} = \mathcal{L}_{\mathrm{MSE}} + \sum_{i=1}^{4} \left( \frac{1}{2\sigma_i^2} \mathcal{L}_{\mathrm{aux}}^{(i)} + \log(\sigma_i) \right),
\label{eq:adf_loss}
\end{equation}
where the primary prediction loss $\mathcal{L}_{\mathrm{MSE}} = \text{MSE}(\hat{\mathbf{Y}}, \mathbf{Y})$ serves as the stable anchor with a fixed unit weight to guarantee magnitude fidelity. To enforce shape alignment with complementary constraints, the regularization term $\mathcal{L}_{\mathrm{aux}}^{(i)}$ integrates four auxiliary objectives:
\textit{1)} $\mathcal{L}_{\mathrm{L1}} = \| \hat{\mathbf{U}}_k \hat{\mathbf{\Sigma}}_k \hat{\mathbf{V}}_k^\top - \mathbf{U}_k \mathbf{\Sigma}_k \mathbf{V}_k^\top \|_1$ retains the top-$k$ singular components to filter out high-frequency noise, forcing the model to align with the global trend; 
\textit{2)} $\mathcal{L}_{\mathrm{Mean}} = \frac{1}{N} \textstyle\sum_{j=1}^{N} ( \mu(\hat{\mathbf{y}}_j) - \mu(\mathbf{y}_j) )^2$ calculates the patch-wise mean to prevent local vertical shifts; 
\textit{3)} $\mathcal{L}_{\mathrm{Var}} = \frac{1}{N} \textstyle\sum_{j=1}^{N} ( \text{std}(\hat{\mathbf{y}}_j) - \text{std}(\mathbf{y}_j) )^2$ uses variance to measure the variation severity, forcing the model to restore severe fluctuations; and 
\textit{4)} $\mathcal{L}_{\mathrm{Corr}} = \frac{1}{N} \textstyle\sum_{j=1}^{N} ( 1 - \frac{\mathrm{Cov}(\hat{\mathbf{y}}_j, \mathbf{y}_j)}{\text{std}(\hat{\mathbf{y}}_j)\text{std}(\mathbf{y}_j)} )$ uses the Pearson correlation to evaluate local shape similarity, ensuring the model accurately captures turning points.
To balance these different objectives, $\sigma_i$ is a learnable uncertainty parameter that automatically down-weights noisy tasks.

During the optimization process guided by $\mathcal{L}_{\mathrm{Total}}$, the pre-trained language encoder and the language embedding layer are kept strictly frozen. Only the following lightweight components are optimized during fine-tuning: \textit{1) Temporal representation learner:} including patch projection, historical auto-correlation modeling, and the linear head; \textit{2) Textual representation learner:} including the cross-attention mechanism and linear heads; \textit{3) Symbolic representation learner:} including the cross-attention mechanisms and the feature mixing module; \textit{4) Multimodal prediction fusion:} including the linear projection layer for VAT routing.


\section{Experiment}
We conduct extensive experiments on 8 public time series datasets, comparing STaT against 9 representative baselines. 
\subsection{Setup}
\textbf{Datasets.} As summarized in Table~\ref{tab:dataset_stats}, we evaluate STaT on eight widely used time series datasets across diverse domains, including: temperature monitoring (ETTm1, ETTm2, ETTh1, ETTh2), electricity consumption (Electricity), transportation (Traffic), weather forecasting (Weather) and daily exchange rates (Exchange) \cite{zhou2021informer, lai2018modeling}. These datasets are commonly used for benchmarking forecasting models \cite{wu2022timesnet}, and vary in frequency, dimensionality, and temporal characteristics. 

\textbf{Metrics.} Performance is measured using magnitude metrics, including MSE and MAE, as well as shape-aware metrics, including Dynamic Time Warping (DTW) \cite{muller2007information} and Time Distortion Index (TDI) \cite{frias2016introducing, le2019shape}, all following standard evaluation practices in this field. The formulations for these two shape-aware metrics are defined as follows:
\begin{equation}
\mathrm{DTW}(\mathbf{Y},\hat{\mathbf{Y}}) = \min_{A\in\mathcal{A}(\mathbf{Y},\hat{\mathbf{Y}})}\sum_{(m,n)\in A} d(\mathbf{y}_m,\hat{\mathbf{y}}_n),
\label{eq:dtw}
\end{equation}
\begin{equation}
\mathrm{TDI}(\mathbf{Y},\hat{\mathbf{Y}}) = \sum_{(m,n)\in A^*}\frac{(m-n)^2}{T^2},
\label{eq:tdi}
\end{equation}
where $(m,n)$ denotes aligned time steps in a warping path $A$ from the admissible set $\mathcal{A}$, $A^*$ is the optimal path minimizing the alignment distance, and $d(\cdot,\cdot)$ is the squared Euclidean distance.

\begin{table}[t]
\centering
\caption{Summary of benchmark datasets. Each dataset includes multiple time series (Dim.) with varying sequence lengths, split into training, validation, and testing sets. Data are collected at different frequencies across various domains.}
\label{tab:dataset_stats}
\renewcommand{\arraystretch}{1.18}
\resizebox{\columnwidth}{!}{%
\begin{tabular}{l c c c c c}
\toprule
\textbf{Dataset} & \textbf{Dim.} & \textbf{Series Length} & \textbf{Dataset Size} & \textbf{Frequency} & \textbf{Domain} \\
\midrule
ETTh1       & 7   & \{96, 192, 336, 720\} & (8545, 2881, 2881)    & 1 hour & Temperature \\
ETTh2       & 7   & \{96, 192, 336, 720\} & (8545, 2881, 2881)    & 1 hour & Temperature \\
ETTm1       & 7   & \{96, 192, 336, 720\} & (34465, 11521, 11521) & 15 min & Temperature \\
ETTm2       & 7   & \{96, 192, 336, 720\} & (34465, 11521, 11521) & 15 min & Temperature \\
Weather     & 21  & \{96, 192, 336, 720\} & (36792, 5271, 10540)  & 10 min & Weather \\
Electricity & 321 & \{96, 192, 336, 720\} & (18317, 2633, 5261)   & 1 hour & Electricity \\
Traffic     & 862 & \{96, 192, 336, 720\} & (12185, 1757, 3509)   & 1 hour & Transportation \\
Exchange    & 8   & \{96, 192, 336, 720\} & (5120, 665, 1422)     & 1 day  & Exchange rates \\
\bottomrule
\end{tabular}%
}
\end{table}

\textbf{Baselines.} We compare STaT with state-of-the-art time series models from three categories. \textit{1) LLM / multimodal time-series models:} TimeCMA \cite{liu2025timecma}, Time-VLM \cite{zhong2025time}, Time-LLM \cite{jin2024time}, and GPT4TS \cite{zhou2023one}, which leverage large language models and/or multimodal priors for forecasting. \textit{2) Transformer-based forecasting models:} We adopt PatchTST \cite{nie2022time}, Non-stationary Transformer \cite{liu2022non}, and FEDformer \cite{zhou2022fedformer}, which are strong sequence modeling baselines for long-horizon forecasting. \textit{3) CNN/linear forecasting models:} TimesNet \cite{wu2022timesnet} and DLinear \cite{zeng2023transformers} are widely used efficient non-LLM baselines. In contrast, STaT proposed in this paper is the first framework to integrate symbolic modality with textual and temporal modalities for time series forecasting. Performance results for some baselines are cited from \cite{liu2024time} where applicable.

\textbf{Implementation Details.} To ensure a fair comparison, we compare STaT against superior baselines using a unified pipeline under the same configurations as \cite{wu2022timesnet} for magnitude metrics. For shape-aware evaluation, we follow the standard protocol established by \cite{le2019shape}. We employ the pre-trained BERT (bert-base-uncased) \cite{devlin2019bert} as the language encoder for both the textual and symbolic representation learners, and freeze its parameters, optimizing only lightweight task‑specific modules. The look-back window length is set to $L=96$, and the forecasting horizons are $T \in \{96, 192, 336, 720\}$. For VAT routing, $\eta$ is set to 2. For multi-scale fABBA, $tol_{\mathrm{fine}}$, $tol_{\mathrm{mid}}$, and $tol_{\mathrm{coarse}}$ are set to 0.01, 0.10, and 0.50, respectively, corresponding to a tolerance ratio of $1{:}10{:}50$. All models are trained with Adam ($10^{-3}$ initial learning rate, halved per epoch), batch size 32, for up to 10 epochs with early stopping. The experiments are implemented using PyTorch and executed on NVIDIA RTX A6000 48GB GPU.

\subsection{Long-Term Forecasting}
\begin{figure}[htbp]
    \centering
    \begin{subfigure}[b]{0.495\linewidth}
        \centering
        \includegraphics[width=\linewidth]{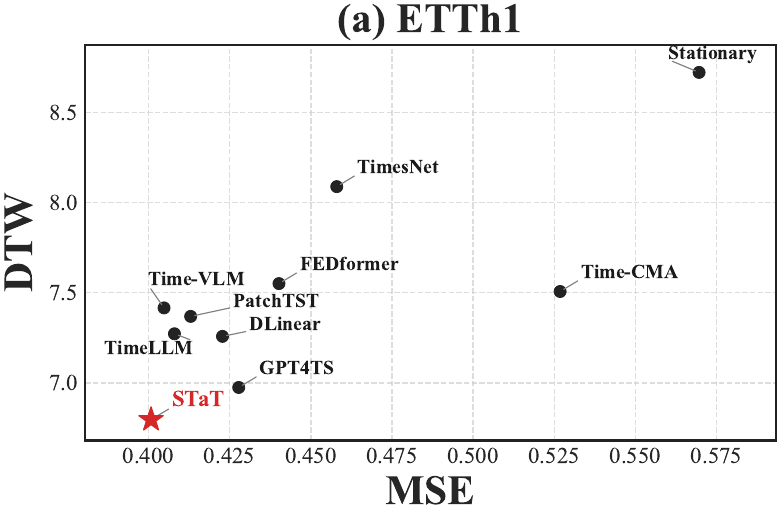}
    \end{subfigure}%
    \hfill
    \begin{subfigure}[b]{0.495\linewidth}
        \centering
        \includegraphics[width=\linewidth]{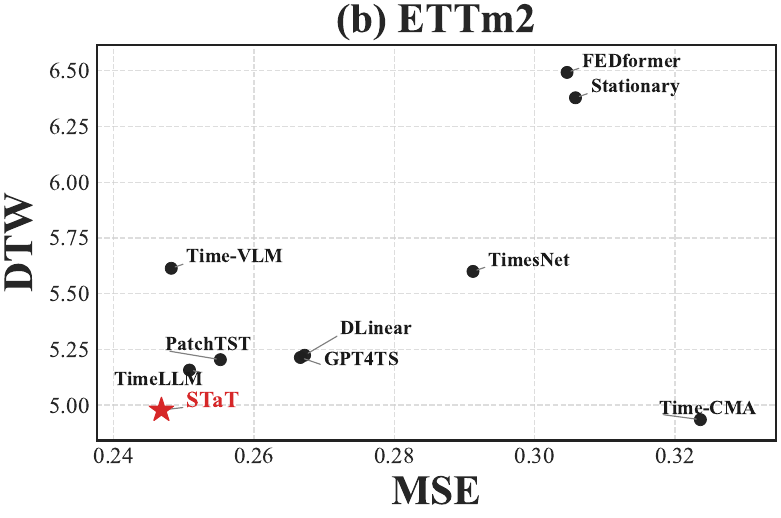}
    \end{subfigure}

    \begin{subfigure}[b]{0.495\linewidth}
        \centering
        \includegraphics[width=\linewidth]{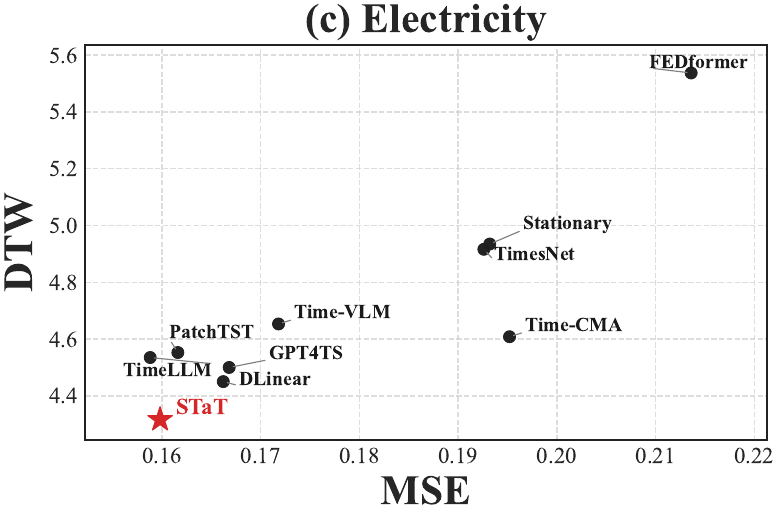}
    \end{subfigure}%
    \hfill
    \begin{subfigure}[b]{0.495\linewidth}
        \centering
        \includegraphics[width=\linewidth]{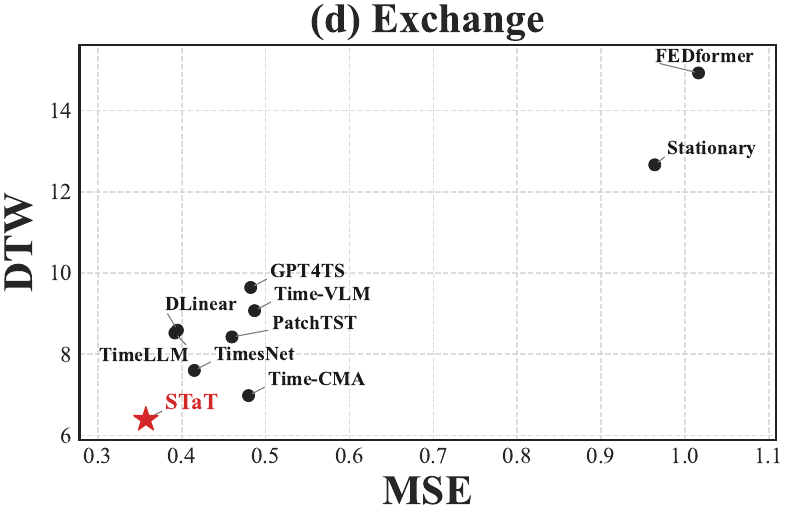}
    \end{subfigure}

    \caption{Comparison of the ability to balance MSE (magnitude error) and DTW (shape error) across different methods.}
    \label{fig:mse_dtw_comparison}
    
\end{figure}

We evaluate the long-term forecasting capabilities of STaT across multiple horizons and datasets. As shown in Table~\ref{tab:main_results_avg}, STaT achieves state-of-the-art overall performance. Notably, on the challenging Exchange dataset (daily foreign-exchange rates with weak periodicity and high non-stationarity), STaT reduces MSE by 8.9\% compared to the second-best model (Time-LLM) and MAE by 2.4\%. It also establishes optimal magnitude accuracy on datasets such as ETTh1, ETTh2, and ETTm2 against strong foundation models. Furthermore, STaT exhibits a pronounced advantage in shape-aware metrics. Specifically, on the Exchange dataset, it significantly reduces the DTW error by 8.5\% compared to the best baseline TimeCMA. Similarly, for the TDI metric, it achieves a 6.9\% error reduction on ETTh2 compared to GPT4TS, an 8.3\% reduction on Traffic, and a 7.1\% reduction on ETTh1 compared to Time-VLM. As further illustrated in Figure~\ref{fig:mse_dtw_comparison}, comparing MSE against DTW across four representative datasets (ETTh1, ETTm2, Electricity, and Exchange) reveals that STaT (red star) consistently occupies the optimal lower-left region, demonstrating that our model successfully achieves a balance in reducing both magnitude errors and shape errors. 

\begin{table*}[htbp]
\centering
\scriptsize
\setlength{\tabcolsep}{5pt}
\caption{Long-term forecasting results. Results are averaged over forecasting horizons $T \in \{96,192,336,720\}$. Lower values indicate better performance. Full results see Appendix~\ref{app:full_long-term_results}. \textbf{Bold}: best, \underline{Underline}: second best.} 

\label{tab:main_results_avg}
\fontsize{10pt}{8pt}\selectfont
\resizebox{\textwidth}{!}{%
\begin{tabular}{ll cccccccccc}
\toprule
\textbf{Dataset} & \textbf{Metric} & \makecell{\textbf{STaT}\\ \normalsize\textbf{(Ours)}} & \makecell{TimeCMA\\ \normalsize{[2025]}} & \makecell{Time-VLM\\ \normalsize{[2025]}} & \makecell{Time-LLM\\ \normalsize{[2024]}} & \makecell{GPT4TS\\ \normalsize{[2023]}} & \makecell{PatchTST\\ \normalsize{[2023]}} & \makecell{TimesNet\\ \normalsize{[2023]}} & \makecell{DLinear\\ \normalsize{[2023]}} & \makecell{Stationary\\ \normalsize{[2022]}} & \makecell{FEDformer\\ \normalsize{[2022]}} \\
\midrule
\multirow{4}{*}{\textbf{ETTh1}}
& MSE & \textbf{0.401} & 0.527 & \underline{0.405} & 0.408 & 0.428 & 0.413 & 0.458 & 0.423 & 0.570 & 0.440 \\

& MAE & \textbf{0.417} & 0.478 & \underline{0.420} & 0.424 & 0.426 & 0.431 & 0.450 & 0.437 & 0.537 & 0.460 \\

& DTW & \textbf{6.797} & 7.507 & 7.416 & 7.273 & \underline{6.975} & 7.369 & 8.088 & 7.258 & 8.721 & 7.551 \\

& TDI & \textbf{0.065} & 0.072 & \underline{0.070} & 0.086 & 0.072 & 0.087 & 0.103 & 0.084 & 0.123 & 0.081 \\

\midrule
\multirow{4}{*}{\textbf{ETTh2}}
& MSE & \textbf{0.327} & 0.461 & 0.341 & 0.334 & 0.355 & \underline{0.330} & 0.414 & 0.431 & 0.526 & 0.437 \\

& MAE & \textbf{0.378} & 0.450 & 0.391 & 0.383 & 0.395 & \underline{0.379} & 0.427 & 0.447 & 0.516 & 0.449 \\

& DTW & \underline{6.303} & 6.532 & 6.763 & 6.778 & \textbf{6.146} & 6.682 & 7.321 & 6.592 & 6.984 & 7.435 \\

& TDI & \textbf{0.121} & 0.135 & 0.135 & 0.147 & \underline{0.130} & 0.147 & 0.175 & 0.145 & 0.178 & 0.171 \\

\midrule
\multirow{4}{*}{\textbf{ETTm1}}
& MSE & \underline{0.344} & 0.453 & 0.351 & \textbf{0.329} & 0.352 & 0.351 & 0.400 & 0.357 & 0.481 & 0.448 \\

& MAE & \underline{0.373} & 0.431 & 0.376 & \textbf{0.372} & 0.383 & 0.381 & 0.406 & 0.379 & 0.456 & 0.452 \\

& DTW & \textbf{5.807} & 6.135 & 6.263 & \underline{5.813} & 5.873 & 5.887 & 6.600 & 6.046 & 7.040 & 7.176 \\

& TDI & \textbf{0.084} & 0.091 & 0.092 & 0.091 & \underline{0.089} & 0.094 & 0.106 & 0.092 & 0.119 & 0.106 \\

\midrule
\multirow{4}{*}{\textbf{ETTm2}}
& MSE & \textbf{0.247} & 0.324 & \underline{0.248} & 0.251 & 0.267 & 0.255 & 0.291 & 0.267 & 0.306 & 0.305 \\

& MAE & \textbf{0.310} & 0.350 & \underline{0.311} & 0.314 & 0.326 & 0.315 & 0.333 & 0.334 & 0.347 & 0.349 \\

& DTW & \underline{4.978} & \textbf{4.935} & 5.614 & 5.158 & 5.214 & 5.204 & 5.600 & 5.224 & 6.379 & 6.493 \\

& TDI & \textbf{0.118} & \underline{0.118} & 0.132 & 0.124 & 0.128 & 0.125 & 0.139 & 0.123 & 0.166 & 0.145 \\

\midrule
\multirow{4}{*}{\textbf{Electricity}}
& MSE & \underline{0.160} & 0.195 & 0.172 & \textbf{0.159} & 0.167 & 0.162 & 0.193 & 0.166 & 0.193 & 0.214 \\

& MAE & \underline{0.253} & 0.294 & 0.272 & 0.253 & 0.263 & \textbf{0.253} & 0.295 & 0.264 & 0.296 & 0.327 \\

& DTW & \textbf{4.316} & 4.608 & 4.653 & 4.535 & 4.500 & 4.553 & 4.916 & \underline{4.450} & 4.935 & 5.537 \\

& TDI & \textbf{0.015} & 0.017 & 0.016 & \underline{0.015} & 0.017 & 0.016 & 0.018 & 0.017 & 0.019 & 0.020 \\

\midrule
\multirow{4}{*}{\textbf{Traffic}}
& MSE & 0.396 & 0.538 & 0.419 & \textbf{0.388} & 0.414 & \underline{0.391} & 0.620 & 0.434 & 0.624 & 0.610 \\

& MAE & 0.267 & 0.354 & 0.304 & \textbf{0.264} & 0.295 & \underline{0.264} & 0.336 & 0.295 & 0.340 & 0.376 \\

& DTW & \underline{6.742} & \textbf{6.364} & 6.972 & 6.763 & 6.873 & 6.823 & 7.745 & 6.848 & 7.792 & 8.402 \\

& TDI & \textbf{0.011} & 0.014 & 0.012 & \underline{0.012} & \underline{0.012} & 0.012 & 0.013 & 0.012 & 0.013 & 0.014 \\

\midrule
\multirow{4}{*}{\textbf{Weather}}
& MSE & \textbf{0.223} & 0.257 & \underline{0.224} & 0.226 & 0.237 & 0.226 & 0.259 & 0.249 & 0.288 & 0.309 \\

& MAE & \underline{0.261} & 0.281 & 0.263 & \textbf{0.258} & 0.271 & 0.264 & 0.287 & 0.300 & 0.314 & 0.360 \\

& DTW & \textbf{4.598} & 5.262 & 4.898 & 4.908 & \underline{4.708} & 4.920 & 5.290 & 5.101 & 5.199 & 6.351 \\

& TDI & \underline{0.153} & 0.169 & 0.160 & 0.158 & 0.158 & 0.157 & 0.174 & 0.173 & 0.162 & \textbf{0.145} \\

\midrule
\multirow{4}{*}{\textbf{Exchange}}
& MSE & \textbf{0.357} & 0.480 & 0.487 & \underline{0.392} & 0.482 & 0.460 & 0.415 & 0.395 & 0.964 & 1.016 \\

& MAE & \textbf{0.407} & 0.471 & 0.473 & \underline{0.417} & 0.461 & 0.460 & 0.441 & 0.420 & 0.656 & 0.761 \\

& DTW & \textbf{6.389} & \underline{6.979} & 9.072 & 8.523 & 9.643 & 8.424 & 7.599 & 8.592 & 12.666 & 14.930 \\

& TDI & \underline{0.191} & 0.249 & 0.277 & 0.242 & 0.260 & 0.254 & 0.254 & 0.245 & 0.223 & \textbf{0.167} \\

\bottomrule
\end{tabular}%
}
\end{table*}

\subsection{Few-Shot Forecasting}
We evaluate the few-shot forecasting capability of STaT by testing its performance using only 10\% of the training data. This setting assesses how effectively STaT performs under limited task-specific supervision. As shown in Table~\ref{tab:fewshot_results_avg}, STaT consistently outperforms most baselines across datasets. For example, on ETTh1, STaT reduces MSE by 5.1\% compared to the best baseline, Time-VLM. On the Exchange dataset, it surpasses GPT4TS by 6.2\% in MAE and DLinear by 4.4\% in MSE. Furthermore, STaT still exhibits a pronounced advantage in shape-aware forecasting, effectively capturing shape alignments where other models fail under limited supervision. Specifically, on ETTm2, it reduces the DTW error by 3.7\% compared to GPT4TS and TDI by 6.3\% compared to Time-VLM. Similarly, on ETTh1, it reduces TDI by 7.2\% compared to Time-VLM, and on ETTm1, it outperforms GPT4TS by 6.4\% in DTW. These improvements demonstrate the effectiveness of multimodal integration when data is scarce. This performance gain stems from the framework's ability to leverage rich pre-trained semantic knowledge while capturing fine-grained structural trends via piecewise symbolic representations, avoiding representation collapse under limited task-specific supervision.

\begin{table*}[htbp]
\centering
\scriptsize
\setlength{\tabcolsep}{5pt}
\caption{Few-shot learning on 10\% training data. We use the same protocol in Table~\ref{tab:main_results_avg}. Full results see Appendix~\ref{app:full_few-shot_results}.} 

\label{tab:fewshot_results_avg}

\fontsize{10pt}{8pt}\selectfont

\resizebox{\textwidth}{!}{%
\begin{tabular}{ll cccccccccc}
\toprule
\textbf{Dataset} & \textbf{Metric} & \makecell{\textbf{STaT}\\ \normalsize\textbf{(Ours)}} & \makecell{TimeCMA\\ \normalsize{[2025]}} & \makecell{Time-VLM\\ \normalsize{[2025]}} & \makecell{Time-LLM\\ \normalsize{[2024]}} & \makecell{GPT4TS\\ \normalsize{[2023]}} & \makecell{PatchTST\\ \normalsize{[2023]}} & \makecell{TimesNet\\ \normalsize{[2023]}} & \makecell{DLinear\\ \normalsize{[2023]}} & \makecell{Stationary\\ \normalsize{[2022]}} & \makecell{FEDformer\\ \normalsize{[2022]}} \\
\midrule
\multirow{4}{*}{\textbf{ETTh1}}
& MSE & \textbf{0.410} & 0.498 & \underline{0.432} & 0.555 & 0.590 & 0.633 & 0.869 & 0.691 & 0.915 & 0.639 \\

& MAE & \textbf{0.428} & 0.470 & \underline{0.442} & 0.522 & 0.525 & 0.542 & 0.628 & 0.600 & 0.639 & 0.561 \\

& DTW & \textbf{6.908} & 7.453 & 7.409 & 9.286 & 7.248 & \underline{7.221} & 7.970 & 7.340 & 8.863 & 8.006 \\

& TDI & \textbf{0.064} & \underline{0.064} & 0.069 & 0.116 & 0.070 & 0.073 & 0.088 & 0.083 & 0.118 & 0.091 \\

\midrule 
\multirow{4}{*}{\textbf{ETTh2}}
& MSE & \textbf{0.348} & 0.432 & \underline{0.361} & 0.371 & 0.397 & 0.415 & 0.479 & 0.605 & 0.462 & 0.466 \\

& MAE & \textbf{0.392} & 0.434 & 0.406 & \underline{0.394} & 0.421 & 0.431 & 0.465 & 0.538 & 0.455 & 0.475 \\

& DTW & \underline{6.259} & 6.639 & 7.507 & 6.897 & 6.804 & \textbf{6.224} & 8.110 & 6.710 & 7.253 & 7.323 \\

& TDI & \textbf{0.128} & \underline{0.131} & 0.169 & 0.149 & 0.145 & 0.140 & 0.185 & 0.157 & 0.186 & 0.178 \\

\midrule
\multirow{4}{*}{\textbf{ETTm1}}
& MSE & \textbf{0.353} & 0.435 & \underline{0.361} & 0.404 & 0.464 & 0.501 & 0.677 & 0.411 & 0.797 & 0.722 \\

& MAE & \textbf{0.381} & 0.428 & \underline{0.382} & 0.427 & 0.441 & 0.466 & 0.537 & 0.429 & 0.578 & 0.605 \\

& DTW & \textbf{5.613} & 6.349 & 6.144 & 8.013 & \underline{5.998} & 6.140 & 6.587 & 6.010 & 7.321 & 7.304 \\

& TDI & \textbf{0.085} & 0.094 & 0.090 & 0.127 & \underline{0.089} & 0.092 & 0.092 & 0.090 & 0.117 & 0.105 \\

\midrule
\multirow{4}{*}{\textbf{ETTm2}}
& MSE & \textbf{0.254} & 0.302 & \underline{0.263} & 0.277 & 0.293 & 0.296 & 0.320 & 0.316 & 0.332 & 0.463 \\

& MAE & \textbf{0.315} & 0.339 & \underline{0.323} & \underline{0.323} & 0.335 & 0.343 & 0.353 & 0.368 & 0.366 & 0.488 \\

& DTW & \textbf{5.033} & 5.511 & 5.376 & 5.736 & \underline{5.225} & 5.636 & 7.055 & 5.312 & 6.964 & 6.453 \\

& TDI & \textbf{0.119} & 0.129 & \underline{0.127} & 0.137 & 0.128 & 0.133 & 0.175 & 0.132 & 0.192 & 0.143 \\

\midrule
\multirow{4}{*}{\textbf{Electricity}}
& MSE & 0.177 & 0.207 & 0.188 & \textbf{0.175} & \underline{0.176} & 0.180 & 0.323 & 0.180 & 0.444 & 0.346 \\

& MAE & 0.273 & 0.305 & 0.291 & \underline{0.270} & \textbf{0.269} & 0.273 & 0.392 & 0.280 & 0.480 & 0.427 \\

& DTW & \textbf{4.420} & 4.668 & 5.059 & 4.600 & \underline{4.460} & 4.865 & 5.709 & 4.651 & 5.518 & 5.913 \\

& TDI & \underline{0.016} & 0.017 & 0.017 & \underline{0.016} & \textbf{0.016} & \underline{0.016} & 0.023 & 0.018 & 0.022 & 0.024 \\

\midrule
\multirow{4}{*}{\textbf{Traffic}}
& MSE & 0.435 & 0.559 & 0.484 & \textbf{0.429} & 0.440 & \underline{0.430} & 0.951 & 0.447 & 1.453 & 0.663 \\

& MAE & 0.311 & 0.383 & 0.357 & \underline{0.306} & 0.310 & \textbf{0.305} & 0.535 & 0.313 & 0.815 & 0.425 \\

& DTW & \underline{6.870} & \textbf{6.657} & 7.861 & 6.925 & 7.218 & 6.960 & 8.817 & 7.276 & 8.158 & 9.334 \\

& TDI & 0.013 & 0.016 & 0.013 & \underline{0.012} & \underline{0.012} & \underline{0.012} & 0.014 & \textbf{0.012} & 0.014 & 0.016 \\

\midrule
\multirow{4}{*}{\textbf{Weather}}
& MSE & \textbf{0.229} & 0.258 & 0.245 & \underline{0.234} & 0.238 & 0.242 & 0.279 & 0.241 & 0.318 & 0.284 \\

& MAE & \textbf{0.266} & 0.284 & 0.282 & \underline{0.273} & 0.275 & 0.279 & 0.301 & 0.283 & 0.323 & 0.324 \\

& DTW & \textbf{4.727} & 5.377 & 4.961 & \underline{4.845} & 4.983 & 4.902 & 5.151 & 5.068 & 5.154 & 6.534 \\

& TDI & \underline{0.155} & 0.168 & 0.162 & 0.158 & 0.159 & 0.159 & 0.161 & 0.171 & 0.169 & \textbf{0.149} \\

\midrule
\multirow{4}{*}{\textbf{Exchange}}
& MSE & \textbf{0.392} & 0.500 & 0.524 & 0.422 & 0.456 & 0.521 & 0.917 & \underline{0.410} & 0.692 & 0.935 \\

& MAE & \textbf{0.427} & 0.477 & 0.475 & 0.482 & \underline{0.455} & 0.468 & 0.685 & 0.470 & 0.574 & 0.724 \\

& DTW & \underline{7.049} & 7.856 & 8.375 & 7.752 & 9.646 & 8.358 & 12.434 & \textbf{6.650} & 10.839 & 13.615 \\

& TDI & \underline{0.209} & 0.269 & 0.291 & 0.255 & 0.251 & 0.276 & 0.236 & 0.249 & 0.234 & \textbf{0.203} \\

\bottomrule
\end{tabular}%
}
\end{table*}

\subsection{Zero-shot Forecasting}
We evaluate the zero-shot forecasting capabilities of STaT under cross-domain scenarios, where the model is tasked with predicting on unseen datasets by effectively transferring knowledge from unrelated domains. To ensure a fair comparison, we follow the previous zero-shot setup~\cite{jin2024time} by using the ETT datasets as source and target domains.

As shown in Table~\ref{tab:cross_domain_avg_results}, STaT demonstrates strong generalizability, outperforming or matching baselines across various transfer settings. For example, in the ETTh1 $\rightarrow$ ETTh2 transfer setting, STaT achieves a 4.2\% lower MSE than Time-LLM and a 1.0\% lower MAE than Time-VLM. In ETTh2 $\rightarrow$ ETTm2, it reduces DTW by 10.6\% compared to PatchTST and TDI by 5.8\% compared to Time-LLM. In ETTm2 $\rightarrow$ ETTh2, STaT not only achieves a 12.0\% lower TDI than DLinear and a 1.8\% lower DTW than GPT4TS, but also performs competitively in magnitude metrics, closely matching Time-LLM with only a 1.4\% difference in MSE and Time-VLM with a 0.3\% difference in MAE. These results highlight the ability of STaT to generalize across domains without fine-tuning, leveraging pre-trained textual semantics and piecewise symbolic representations for effective knowledge transfer.

\begin{table}[htbp]
\centering
\scriptsize
\setlength{\tabcolsep}{5pt}
\caption{Zero-shot learning results.}
\label{tab:cross_domain_avg_results}

\fontsize{18pt}{14pt}\selectfont

\resizebox{\columnwidth}{!}{%
\begin{tabular}{llcccccc}
\toprule
\rule{0pt}{3.2ex}
\textbf{Task} & \textbf{Metric} 
& \makecell{\textbf{STaT}\\ \huge\textbf{(Ours)}} 
& \makecell{Time-VLM\\ \huge{[2025]}} 
& \makecell{Time-LLM\\ \huge{[2024]}} 
& \makecell{GPT4TS\\ \huge{[2023]}} 
& \makecell{PatchTST\\ \huge{[2023]}} 
& \makecell{DLinear\\ \huge{[2023]}} \\
\midrule

\multirow{4}{*}{\textbf{ETTh1 $\rightarrow$ ETTh2}}
& MSE & \textbf{0.338} & \textbf{0.338} & \underline{0.353} & 0.406 & 0.380 & 0.493 \\

& MAE & \textbf{0.381} & \underline{0.385} & 0.387 & 0.422 & 0.405 & 0.488 \\

& DTW & \textbf{6.172} & 6.709 & 7.872 & \underline{6.555} & 6.673 & 6.865 \\

& TDI & \textbf{0.130} & 0.136 & 0.162 & \underline{0.132} & 0.148 & 0.161 \\

\midrule

\multirow{4}{*}{\textbf{ETTh1 $\rightarrow$ ETTm2}}
& MSE & \underline{0.282} & 0.293 & \textbf{0.273} & 0.325 & 0.314 & 0.415 \\

& MAE & \underline{0.345} & 0.350 & \textbf{0.340} & 0.363 & 0.360 & 0.452 \\

& DTW & \textbf{5.602} & 6.266 & 7.077 & 6.185 & \underline{5.873} & 6.858 \\

& TDI & \textbf{0.167} & 0.182 & 0.187 & 0.180 & \underline{0.180} & 0.222 \\

\midrule

\multirow{4}{*}{\textbf{ETTh2 $\rightarrow$ ETTh1}}
& MSE & \underline{0.492} & 0.496 & \textbf{0.479} & 0.757 & 0.565 & 0.703 \\

& MAE & \textbf{0.472} & 0.480 & \underline{0.474} & 0.578 & 0.513 & 0.574 \\

& DTW & \textbf{7.402} & 8.376 & 8.932 & 8.334 & 8.601 & \underline{7.558} \\

& TDI & \textbf{0.078} & 0.089 & 0.098 & 0.086 & 0.108 & \underline{0.084} \\

\midrule

\multirow{4}{*}{\textbf{ETTh2 $\rightarrow$ ETTm2}}
& MSE & \underline{0.278} & 0.297 & \textbf{0.272} & 0.335 & 0.325 & 0.328 \\

& MAE & \underline{0.341} & 0.353 & \textbf{0.341} & 0.370 & 0.365 & 0.386 \\

& DTW & \textbf{5.374} & 6.134 & 6.148 & 6.367 & \underline{6.010} & 6.126 \\

& TDI & \textbf{0.162} & 0.181 & \underline{0.172} & 0.181 & 0.173 & 0.180 \\

\midrule

\multirow{4}{*}{\textbf{ETTm1 $\rightarrow$ ETTh2}}
& MSE & \underline{0.365} & \textbf{0.354} & 0.381 & 0.433 & 0.439 & 0.464 \\

& MAE & \underline{0.405} & \textbf{0.397} & 0.412 & 0.439 & 0.438 & 0.475 \\

& DTW & \underline{6.375} & 6.904 & 7.287 & 6.535 & \textbf{6.320} & 6.713 \\

& TDI & \textbf{0.143} & 0.158 & 0.154 & \underline{0.147} & 0.150 & 0.150 \\

\midrule

\multirow{4}{*}{\textbf{ETTm1 $\rightarrow$ ETTm2}}
& MSE & \textbf{0.263} & \underline{0.264} & 0.268 & 0.313 & 0.296 & 0.335 \\

& MAE & \underline{0.319} & \textbf{0.319} & 0.320 & 0.348 & 0.334 & 0.389 \\

& DTW & \underline{5.063} & 5.408 & 5.734 & 5.164 & \textbf{4.947} & 5.509 \\

& TDI & \underline{0.122} & 0.126 & 0.125 & \textbf{0.120} & 0.127 & 0.137 \\

\midrule

\multirow{4}{*}{\textbf{ETTm2 $\rightarrow$ ETTh2}}
& MSE & \underline{0.359} & 0.359 & \textbf{0.354} & 0.435 & 0.409 & 0.455 \\

& MAE & \underline{0.400} & \textbf{0.399} & \underline{0.400} & 0.443 & 0.425 & 0.471 \\

& DTW & \textbf{6.643} & 7.121 & 7.643 & \underline{6.765} & 7.323 & 6.889 \\

& TDI & \textbf{0.139} & 0.174 & 0.178 & \underline{0.158} & 0.164 & \underline{0.158} \\

\midrule

\multirow{4}{*}{\textbf{ETTm2 $\rightarrow$ ETTm1}}
& MSE & \textbf{0.409} & 0.432 & \underline{0.414} & 0.485 & 0.568 & 0.649 \\

& MAE & \textbf{0.423} & \underline{0.426} & 0.438 & 0.454 & 0.492 & 0.537 \\

& DTW & \textbf{6.898} & 7.727 & 7.862 & \underline{7.087} & 7.707 & 7.270 \\

& TDI & \textbf{0.104} & 0.113 & \underline{0.105} & 0.109 & 0.113 & 0.108 \\

\bottomrule
\end{tabular}%
}
\end{table}

\subsection{Model Analysis}
\label{sec:model_analysis}
\textbf{Ablation Studies.} Table~\ref{tab:ablation_multirow} evaluates the contributions of key components of STaT, including \textit{textual representation learner} (denoted by TRL), \textit{symbolic representation learner} (denoted by SRL), \textit{VAT routing mechanism}, and \textit{ADF Loss}. Note that the temporal representation learner is not evaluated separately, as it serves as the query module for the other two modalities and cannot be removed independently. 

From Table~\ref{tab:ablation_multirow}, the full model achieves optimal performance for both MSE and DTW. Notably, the SRL's contribution is the most significant as excluding it leads to the largest performance drop, with an average MSE increase of 6.48\% and DTW of 13.92\% , demonstrating the necessity of symbolic time series analysis for precise shape alignment. Excluding the ADF Loss also results in a significant degradation in shape fidelity, with DTW increasing by 9.06\% , confirming its role in providing explicit optimization guidance for shape preservation. Even when optimized solely with MSE, the \textit{w/o ADF Loss} variant (DTW: 7.413) still outperforms the strong baseline Time-VLM (DTW: 7.416), confirming that the symbolic modality inherently contributes to precise shape alignment. Removing the VAT module results in an average MSE increase of 5.74\% , highlighting its role in dynamically routing and balancing the multimodalities to stabilize magnitude predictions. Finally, excluding the TRL results in a more modest degradation, with an average MSE and DTW increase of 2.74\% and 4.34\% respectively, validating its ability to provide useful global semantic context.
\begin{table}[htbp]
\centering
\scriptsize
\setlength{\tabcolsep}{0pt}
\caption{Ablation study on multimodal components over forecasting horizons $T \in \{96, 192, 336, 720\}$ on ETTh1 dataset, with both MSE and DTW performance degradation (\%Deg) measured for each variant. \textbf{Bold}: best, \underline{Underline}: second best.}
\label{tab:ablation_multirow}
\fontsize{8pt}{6pt}\selectfont
\begin{tabular*}{\columnwidth}{@{\extracolsep{\fill}} ccccccc @{}}
\toprule
\textbf{Horizon} & \textbf{Metric} & \textbf{Full} & \textbf{w/o TRL} & \textbf{w/o SRL} & \textbf{w/o VAT} & \textbf{w/o ADF Loss} \\
\midrule
\multirow{2}{*}{96} & MSE & \textbf{0.354} & \underline{0.363} & 0.374 & 0.369 & 0.367 \\
 & DTW & \textbf{3.673} & \underline{3.678} & 3.932 & 3.712 & 3.851 \\
\midrule
\multirow{2}{*}{192} & MSE & \textbf{0.394} & \underline{0.397} & 0.405 & 0.413 & 0.404 \\
 & DTW & \textbf{5.426} & 5.613 & 5.844 & \underline{5.498} & 5.627 \\
\midrule
\multirow{2}{*}{336} & MSE & \textbf{0.417} & \underline{0.432} & 0.439 & 0.433 & 0.433 \\
 & DTW & \textbf{7.177} & 7.361 & 7.986 & \underline{7.291} & 7.507 \\
\midrule
\multirow{2}{*}{720} & MSE & \textbf{0.438} & \underline{0.454} & 0.488 & 0.481 & 0.482 \\
 & DTW & \textbf{10.911} & 11.713 & 13.211 & \underline{11.352} & 12.668 \\
\midrule
\multirow{2}{*}{Avg} & MSE & \textbf{0.401} & \underline{0.412} & 0.427 & 0.424 & 0.422 \\
 & DTW & \textbf{6.797} & 7.092 & 7.743 & \underline{6.963} & 7.413 \\
\midrule
\multirow{2}{*}{\%Deg} & MSE & - & 2.74\% $\scriptstyle\uparrow$ & 6.48\% $\scriptstyle\uparrow$ & 5.74\% $\scriptstyle\uparrow$ & 5.24\% $\scriptstyle\uparrow$ \\
 & DTW & - & 4.34\% $\scriptstyle\uparrow$ & 13.92\% $\scriptstyle\uparrow$ & 2.44\% $\scriptstyle\uparrow$ & 9.06\% $\scriptstyle\uparrow$ \\
\bottomrule
\end{tabular*}
\end{table}

\begin{figure}[htbp]
\centering
\IfFileExists{ad4.pdf}{\includegraphics[width=\columnwidth]{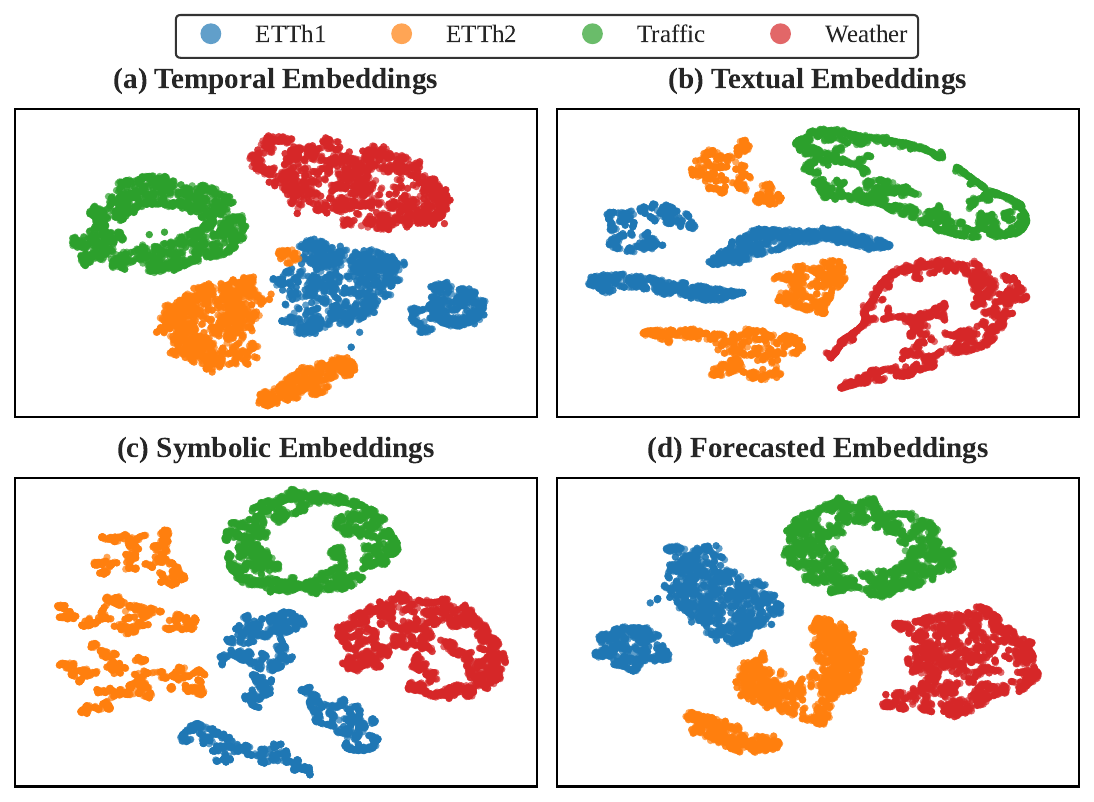}}{\rule{\columnwidth}{4cm}}
\caption{The combined t-SNE visualization of temporal, textual, symbolic, and forecasted embeddings across four datasets.}
\label{fig:tsne}
\end{figure}

\textbf{Interpretability Analysis.}
We employ t-SNE to visualize the learned embeddings across four representation spaces on four datasets, as shown in Figure~\ref{fig:tsne}. Figure~\ref{fig:tsne}(a) illustrates the temporal embeddings, showing a basic clustering of the datasets. This separation indicates the domain-specific continuous patterns across different datasets. Figure~\ref{fig:tsne}(b) visualizes the textual embeddings. The dispersed distribution reflects the discrete semantic nature of text, suggesting the model captures diverse and localized semantic concepts. The embeddings of ETTh1 and ETTh2 overlap semantically, as they originate from datasets with a common semantic domain. Figure~\ref{fig:tsne}(c) shows the symbolic embeddings, which exhibit distinct structural shapes across different datasets. For instance, the Traffic and Weather datasets form clear ring structures, indicating that the symbolic modality can successfully extract periodic properties, even for high-dimensional sequences. Notably, although ETTh1 and ETTh2 display similar structural shapes in the symbolic space, their embeddings hardly overlap, suggesting the symbolic modality can capture dataset-specific structural characteristics. Finally, Figure~\ref{fig:tsne}(d) displays the forecasted embeddings prior to the final prediction. The clusters here exhibit tighter cohesion and clearer boundaries. This enhanced discriminability suggests the model learns effective representations by synthesizing complementary information from all modalities. Overall, these visualizations demonstrate the ability of STaT to integrate temporal trends, textual semantics, and symbolic shapes into comprehensive multimodal representations.

\begin{figure}[htbp]
\centering
\IfFileExists{ac.pdf}{\includegraphics[width=\columnwidth]{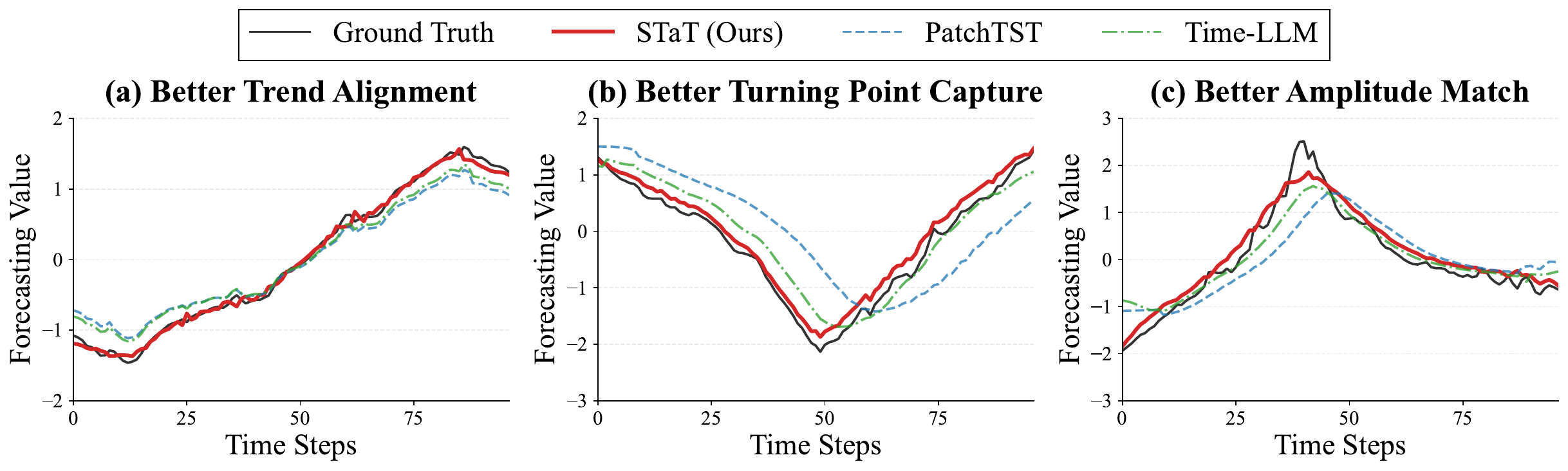}}{\rule{\columnwidth}{4cm}}
\caption{Forecasting visualization on the Weather dataset. We compare STaT with representative baselines (PatchTST and Time-LLM) in terms of (a) trend alignment, (b) turning point capture, and (c) amplitude match.}
\label{fig:forecasting_vis}
\end{figure}

\textbf{Forecasting Visualization.}
We visualize the forecasting results on the Weather dataset. Compared to representative baselines (PatchTST and Time-LLM), STaT demonstrates superior forecasting performance, as shown in Figure~\ref{fig:forecasting_vis}. Specifically: (a) our predictions follow the actual trajectory more closely, achieving the optimal trend alignment; (b) our predictions respond to sharp drops with reduced phase shifts, locating the valleys more accurately on the time axis, demonstrating superior turning point capture; and (c) our curve aligns more closely with the actual values along the vertical axis, resulting in an improved amplitude match. These results indicate that STaT achieves superior overall shape alignment.

\begin{figure}[htbp]
\centering
\begin{subfigure}[b]{0.48\columnwidth}
\centering
\IfFileExists{a.pdf}{\includegraphics[width=\linewidth]{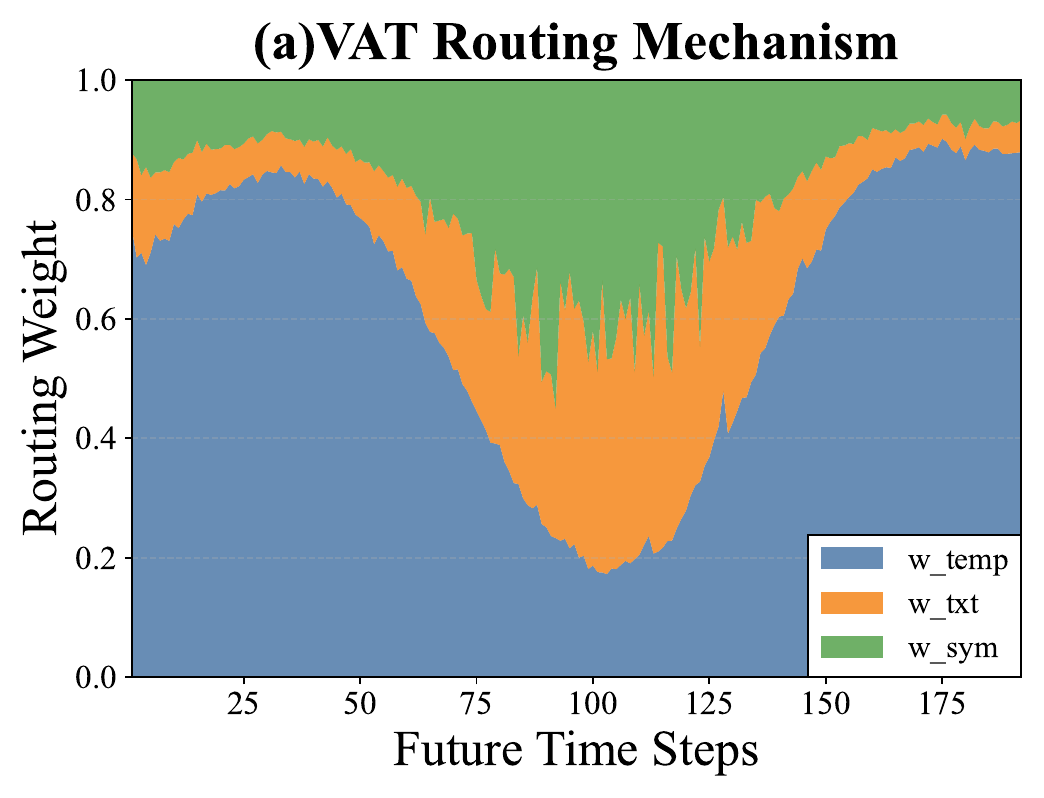}}{\rule{\linewidth}{4cm}}
\end{subfigure}%
\hfill%
\begin{subfigure}[b]{0.48\columnwidth}
\centering
\IfFileExists{b.pdf}{\includegraphics[width=\linewidth]{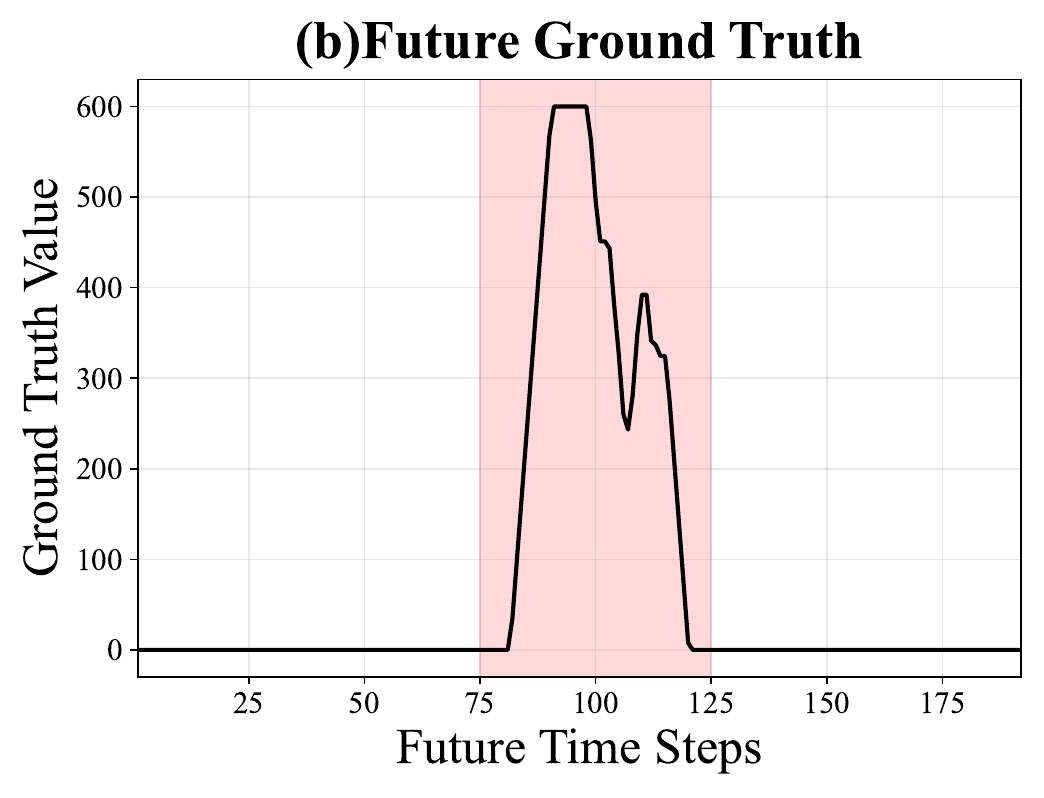}}{\rule{\linewidth}{4cm}}
\end{subfigure}
\caption{Analysis of VAT routing mechanism. (a) The evolution of VAT routing weights ($w_{\mathrm{temp}}$, $w_{\mathrm{txt}}$, $w_{\mathrm{sym}}$) over future time steps. (b) The corresponding ground truth values.}
\label{fig:mechanism}
\end{figure}

\textbf{Analysis of VAT Routing Mechanism.}
To understand the VAT routing mechanism of STaT, Figure~\ref{fig:mechanism} analyzes the adaptive variation of the three modality weights ($w_{\mathrm{temp}}$, $w_{\mathrm{txt}}$, $w_{\mathrm{sym}}$) under the VAT routing mechanism on the Weather dataset with a prediction horizon of $T=192$. Specifically, Figure~\ref{fig:mechanism}(a) illustrates the evolution of the three modality routing weights over the same horizon, and Figure~\ref{fig:mechanism}(b) plots the corresponding future ground truth values, highlighting a period of drastic fluctuation and structural change (shaded in red) between time steps~75 and~125. During smooth and stable periods (e.g., before step 75 and after step 125), the temporal weight ($w_{\mathrm{temp}}$) heavily dominates the prediction, relying on historical auto-correlation for stable trend continuation. Conversely, when encountering the sudden structural spike in the ground truth, the VAT mechanism dynamically captures the high volatility and significantly expands the routing weights for the textual ($w_{\mathrm{txt}}$) and symbolic ($w_{\mathrm{sym}}$) modalities. This demonstrates that our VAT mechanism effectively handles non-stationary environments by adaptively adjusting the weights among different learners based on volatility.

\begin{table}[htbp]
\centering
\scriptsize
\setlength{\tabcolsep}{0pt}
\caption{Computational efficiency comparison between STaT and Time-LLM across datasets. "-" denotes memory exceeds 48GB, making it infeasible on a single GPU. Results are averaged over four prediction lengths under consistent conditions. \textbf{Bold}: best, \underline{Underline}: second best.}
\label{tab:computation}
\fontsize{7pt}{4pt}\selectfont
\begin{tabular*}{\columnwidth}{@{\extracolsep{\fill}} clccccc @{}}
\toprule
Method & Metric & ETTh1 & ETTm2 & Weather & Electricity & Traffic \\
\midrule
\multirow{3}{*}{STaT} & Param. (M)     & \textbf{65.7} & \textbf{65.7} & \textbf{65.7} & \textbf{65.7} & \textbf{65.7} \\
 & Mem. (MB)      & \textbf{673} & \textbf{672} & \textbf{730} & \textbf{1923} & \textbf{4072} \\
 & Speed (s/iter) & \textbf{0.045}   & \textbf{0.038}   & \textbf{0.045}   & \textbf{0.051}    & \textbf{0.097}   \\
\midrule
\multirow{3}{*}{Time-LLM} & Param. (M)     & \underline{3404.6} & \underline{3404.6} & - & - & - \\
 & Mem. (MB)      & \underline{37723} & \underline{37849} & - & - & - \\
 & Speed (s/iter) & \underline{0.607}   & \underline{0.265}   & -   & - & - \\
\bottomrule
\end{tabular*}
\end{table}

\textbf{Computation Studies.}
Table~\ref{tab:computation} compares the computational resources of STaT and Time-LLM across different datasets. We adopt the default \textit{LLaMA-7B} for the Time-LLM baseline. From Table~\ref{tab:computation}, despite the inclusion of multimodal processing, STaT requires only 65.7M parameters, which is orders of magnitude fewer than the baseline. STaT's memory usage scales from 672 MB (ETTm2) to 4072 MB (Traffic), naturally adapting to dataset complexity. Inference speed ranges from 0.038 s/iter (ETTm2) to approximately 0.097 s/iter (Traffic), efficiently handling varying computational loads. In contrast, Time-LLM requires substantially more memory and longer inference times even on smaller datasets like ETTh1 and ETTm2 (e.g., 37723 MB and 0.607 s/iter). This is mainly because Time-LLM utilizes the massive LLaMA-7B backbone as its core forecasting engine, whereas STaT offloads the actual forecasting workload to lightweight temporal and symbolic learners, using the frozen language encoder solely for semantic extraction. Furthermore, for high-dimensional datasets such as Electricity and Traffic, Time-LLM's memory consumption exceeds 48GB, making it infeasible on a single GPU. These results highlight STaT's lightweight design, spatial efficiency, and practical scalability for complex multivariate forecasting.
\begin{figure}[htbp]
\centering
\IfFileExists{Hyperparameter.pdf}{\includegraphics[width=\columnwidth]{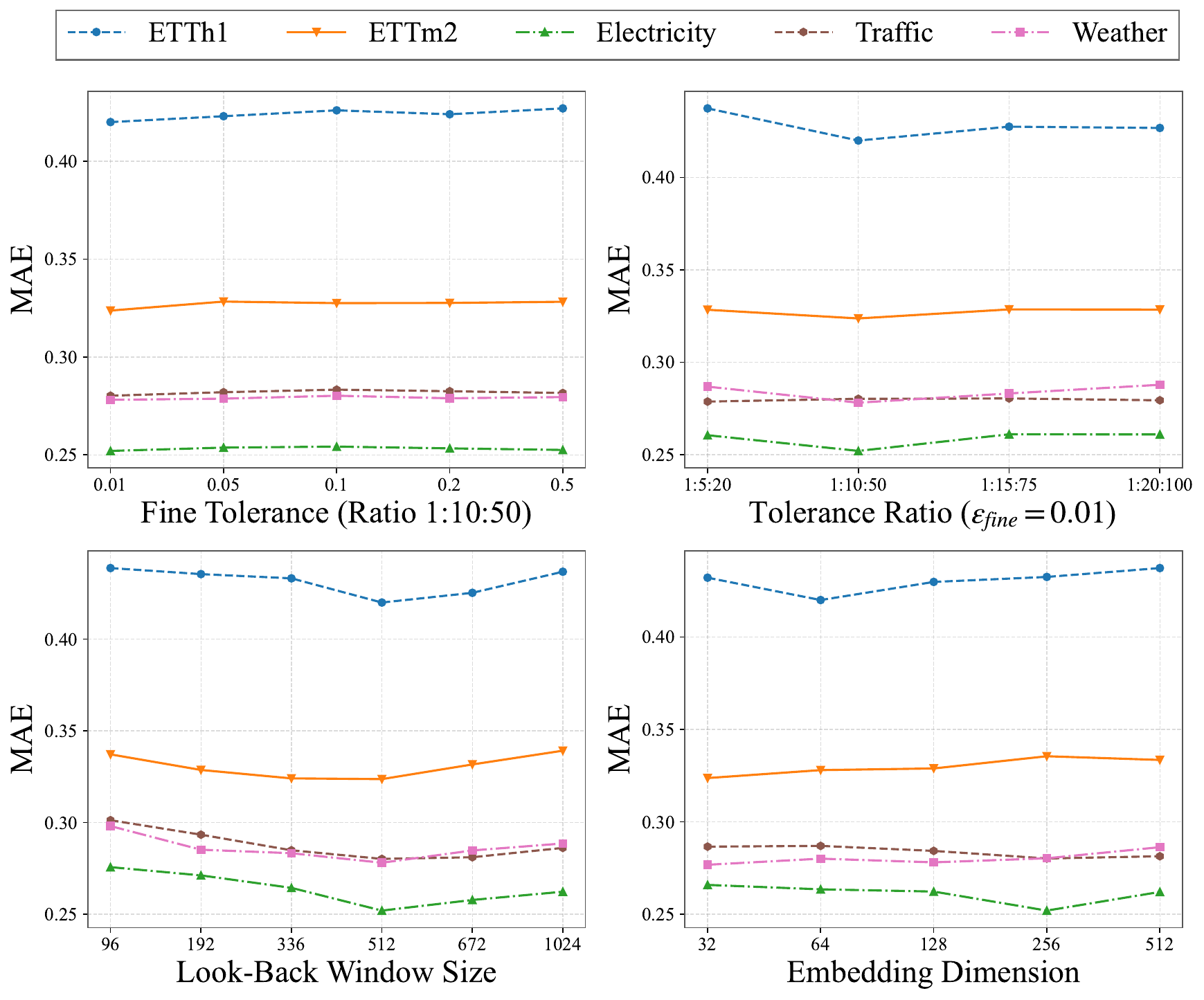}}{\rule{\columnwidth}{5cm}}
\caption{Hyperparameter sensitivity analysis on fABBA tolerance, fABBA tolerance ratio, look-back window size and embedding dimension.}
\label{fig:hparam_mae}
\end{figure}

\textbf{Hyperparameter Sensitivity.} We analyze the impact of key hyperparameters on performance, including \textit{fABBA tolerance},  \textit{fABBA tolerance ratio}, \textit{look-back window size}, and  \textit{embedding dimension}, as shown in Figure~\ref{fig:hparam_mae}. The  \textit{fABBA tolerance} and  \textit{fABBA tolerance ratio} determine the granularity of multi-scale symbolic discretization (Eqn.~(\ref{eq:multiscale_symbol})), while  \textit{Look-back window size} controls the available historical context, and  \textit{embedding dimension} controls the capacity of the latent representation space. For \textit{fABBA tolerance}, performance peaks at $tol_{fine}=0.01$ across all five datasets. Since Z-score normalization makes the tolerance act on relative fluctuations, larger tolerances over-smooth symbolic sequences and weaken local pattern differences. For the \textit{fABBA tolerance ratio}, 1:10:50 gives the best average MAE on four of the five datasets, indicating a good balance across fine, mid, and coarse scales. Smaller ratios limit scale diversity, whereas larger ratios weaken sensitivity to mid-range temporal patterns. \textit{Look-back window size} performs best in the tested range of 96 to 1024, with 512 being optimal for most datasets. Longer window size introduce noise without clear gains, suggesting local temporal patterns are sufficient for accurate forecasting. \textit{Embedding dimension} shows dataset-dependent behavior: $D_{\text{model}}=64\text{--}128$ is sufficient for ETTm2 and Weather, whereas Electricity and Traffic benefit from larger dimensions up to 256 due to more complex cross-variate dynamics. Overall, although each hyperparameter has a certain impact, the four groups of curves stay relatively flat, confirming STaT is robust to hyperparameter variations.

\section{Conclusions}
In this paper, we propose STaT, a novel framework that integrates the symbolic, textual, and temporal modalities for time series forecasting. Recognizing that conventional forecasting mechanisms often sacrifice the capture of volatility for lower magnitude errors, we pioneer the incorporation of discrete symbolic representations to explicitly capture critical turning points and extreme amplitudes. To fully harness the complementarity of these three modalities, we design a VAT routing mechanism that adaptively balances their fusion weights according to the input’s volatility, and an ADF loss that jointly optimizes magnitude accuracy and shape alignment. Extensive experiments on eight datasets show that STaT achieves state-of-the-art performance, improving both magnitude and shape fidelity while maintaining a lightweight memory footprint.

\clearpage

\appendix
\begin{table*}
\section{Long-Term Forecasting}
\label{app:full_long-term_results}

\centering
\scriptsize
\setlength{\tabcolsep}{2.5pt} 
\renewcommand{\arraystretch}{1.0}
\caption{Full long-term forecasting results with forecasting horizons $T \in \{96,192,336,720\}$. A lower value indicates better performance. \textbf{Bold}: best, \underline{Underline}: second best.} 

\label{tab:main_results_top_conference}

\fontsize{11pt}{12pt}\selectfont

\resizebox{\textwidth}{!}{%
\begin{tabular}{ll cccc cccc cccc cccc cccc cccc cccc cccc}
\toprule
\multirow{2}{*}{\textbf{Models}} & \multirow{2}{*}{\textbf{Metrics}}
& \multicolumn{4}{c}{\textbf{ETTh1}}
& \multicolumn{4}{c}{\textbf{ETTh2}}
& \multicolumn{4}{c}{\textbf{ETTm1}}
& \multicolumn{4}{c}{\textbf{ETTm2}}
& \multicolumn{4}{c}{\textbf{Electricity}}
& \multicolumn{4}{c}{\textbf{Traffic}}
& \multicolumn{4}{c}{\textbf{Weather}}
& \multicolumn{4}{c}{\textbf{Exchange}} \\
\cmidrule(lr){3-6} \cmidrule(lr){7-10} \cmidrule(lr){11-14} \cmidrule(lr){15-18} 
\cmidrule(lr){19-22} \cmidrule(lr){23-26} \cmidrule(lr){27-30} \cmidrule(lr){31-34}
& & \textbf{96} & \textbf{192} & \textbf{336} & \textbf{720} & \textbf{96} & \textbf{192} & \textbf{336} & \textbf{720} & \textbf{96} & \textbf{192} & \textbf{336} & \textbf{720} & \textbf{96} & \textbf{192} & \textbf{336} & \textbf{720} & \textbf{96} & \textbf{192} & \textbf{336} & \textbf{720} & \textbf{96} & \textbf{192} & \textbf{336} & \textbf{720} & \textbf{96} & \textbf{192} & \textbf{336} & \textbf{720} & \textbf{96} & \textbf{192} & \textbf{336} & \textbf{720} \\
\midrule

\multirow{4}{*}{\makecell[l]{\textbf{STaT} \\ \textbf{[Ours]}}}
& MSE & \textbf{0.354} & \textbf{0.394} & \textbf{0.417} & \textbf{0.438} & \textbf{0.265} & \textbf{0.321} & \underline{0.343} & \underline{0.378} & \underline{0.289} & \underline{0.327} & \underline{0.361} & \underline{0.401} & \textbf{0.159} & \textbf{0.214} & \textbf{0.269} & \textbf{0.345} & \underline{0.130} & \textbf{0.151} & \underline{0.162} & \underline{0.196} & 0.369 & 0.382 & 0.396 & 0.435 & \underline{0.148} & \underline{0.192} & \textbf{0.243} & \underline{0.311} & \textbf{0.081} & \textbf{0.167} & \textbf{0.294} & \textbf{0.887} \\
& MAE & \textbf{0.383} & \textbf{0.411} & \textbf{0.420} & \textbf{0.453} & \textbf{0.326} & \textbf{0.372} & \underline{0.393} & \underline{0.421} & \underline{0.341} & \underline{0.363} & \textbf{0.382} & \textbf{0.407} & \textbf{0.248} & \underline{0.291} & \textbf{0.324} & \textbf{0.376} & \textbf{0.222} & \underline{0.241} & \underline{0.252} & \underline{0.297} & 0.252 & 0.258 & 0.268 & 0.291 & \textbf{0.194} & \underline{0.239} & \textbf{0.278} & \underline{0.332} & \textbf{0.205} & \textbf{0.295} & \textbf{0.417} & \textbf{0.712} \\
& DTW & \textbf{3.673} & \textbf{5.426} & \textbf{7.177} & \textbf{10.911} & \textbf{3.041} & \textbf{4.641} & \textbf{6.612} & \underline{10.917} & \underline{2.880} & \underline{4.360} & \underline{6.254} & \underline{9.733} & \underline{2.198} & \underline{3.670} & \underline{5.266} & \textbf{8.778} & \underline{2.246} & \underline{3.363} & \textbf{4.644} & \textbf{7.013} & \underline{3.628} & \underline{5.325} & \textbf{7.130} & \underline{10.886} & \textbf{1.869} & \textbf{3.211} & \textbf{4.876} & \textbf{8.434} & \textbf{1.950} & \textbf{4.313} & \textbf{6.490} & \textbf{12.802} \\
& TDI & \textbf{0.069} & \textbf{0.063} & \textbf{0.068} & \textbf{0.062} & \textbf{0.110} & \textbf{0.123} & \textbf{0.121} & \textbf{0.129} & \textbf{0.109} & \textbf{0.085} & \textbf{0.074} & \textbf{0.069} & \underline{0.115} & \underline{0.112} & \underline{0.112} & \underline{0.132} & \textbf{0.022} & \underline{0.016} & \underline{0.012} & \underline{0.011} & \underline{0.016} & \textbf{0.013} & \textbf{0.010} & \textbf{0.006} & \textbf{0.158} & \underline{0.149} & \underline{0.151} & \underline{0.156} & \textbf{0.148} & \underline{0.221} & \underline{0.195} & \underline{0.201} \\
\specialrule{\lightrulewidth}{1pt}{1pt}

\multirow{4}{*}{\makecell[l]{TimeCMA \\ {[2025]}}}
& MSE & 0.448 & 0.502 & 0.535 & 0.622 & 0.335 & 0.468 & 0.512 & 0.528 & 0.356 & 0.440 & 0.491 & 0.523 & 0.205 & 0.282 & 0.344 & 0.463 & 0.165 & 0.187 & 0.196 & 0.233 & 0.519 & 0.525 & 0.543 & 0.565 & 0.160 & 0.217 & 0.279 & 0.372 & 0.125 & 0.237 & 0.477 & 1.079 \\
& MAE & 0.433 & 0.458 & 0.487 & 0.533 & 0.375 & 0.445 & 0.479 & 0.501 & 0.380 & 0.422 & 0.451 & 0.472 & 0.279 & 0.327 & 0.366 & 0.429 & 0.265 & 0.289 & 0.296 & 0.325 & 0.346 & 0.350 & 0.350 & 0.369 & 0.204 & 0.257 & 0.302 & 0.360 & 0.249 & 0.350 & 0.493 & 0.790 \\
& DTW & 3.864 & 5.839 & 8.009 & 12.316 & 3.111 & 5.024 & 7.061 & 10.932 & 2.992 & 4.622 & 6.537 & 10.388 & \textbf{2.197} & \textbf{3.522} & \textbf{5.143} & 8.879 & 2.433 & 3.723 & \underline{4.654} & 7.623 & 3.964 & 5.845 & 7.384 & \textbf{8.263} & 1.985 & 3.587 & 5.604 & 9.871 & 2.344 & 4.477 & 7.838 & \underline{13.257} \\
& TDI & \underline{0.072} & \underline{0.065} & 0.070 & 0.080 & \underline{0.112} & 0.132 & 0.140 & 0.157 & 0.116 & 0.093 & 0.082 & \underline{0.072} & \textbf{0.113} & \textbf{0.110} & 0.119 & \textbf{0.130} & 0.024 & 0.017 & 0.013 & 0.012 & 0.019 & 0.015 & 0.012 & 0.009 & 0.173 & 0.161 & 0.168 & 0.174 & 0.245 & 0.247 & 0.232 & 0.271 \\
\specialrule{\lightrulewidth}{1pt}{1pt}

\multirow{4}{*}{\makecell[l]{Time-VLM \\ {[2025]}}}
& MSE & \underline{0.361} & \underline{0.397} & \underline{0.420} & \underline{0.441} & \underline{0.267} & \underline{0.326} & 0.357 & 0.412 & 0.304 & 0.332 & 0.364 & 0.402 & \underline{0.160} & \underline{0.215} & \underline{0.270} & \underline{0.348} & 0.142 & 0.157 & 0.174 & 0.214 & 0.393 & 0.405 & 0.420 & 0.459 & 0.148 & 0.193 & \textbf{0.243} & 0.312 & 0.103 & 0.224 & 0.463 & 1.157 \\
& MAE & \underline{0.386} & \underline{0.415} & \underline{0.421} & 0.458 & 0.335 & \underline{0.373} & 0.406 & 0.449 & 0.346 & 0.366 & \underline{0.383} & \underline{0.410} & \underline{0.250} & \textbf{0.291} & \underline{0.325} & \underline{0.378} & 0.245 & 0.260 & 0.276 & 0.308 & 0.290 & 0.296 & 0.305 & 0.323 & 0.200 & 0.240 & 0.281 & 0.332 & 0.229 & 0.341 & 0.508 & 0.812 \\
& DTW & 3.786 & 5.648 & 7.772 & 12.458 & 3.208 & 5.203 & 7.255 & 11.388 & 3.182 & 4.815 & 6.623 & 10.432 & 2.585 & 4.065 & 5.908 & 9.898 & 2.350 & 3.484 & 4.893 & 7.886 & 3.784 & 5.540 & 7.319 & 11.247 & 2.057 & 3.441 & 5.276 & 8.821 & 2.292 & 4.739 & 9.371 & 19.887 \\
& TDI & 0.073 & 0.066 & 0.069 & \underline{0.073} & 0.120 & 0.134 & 0.136 & 0.151 & 0.117 & 0.093 & 0.081 & 0.076 & 0.125 & 0.121 & 0.134 & 0.149 & 0.023 & 0.016 & 0.013 & 0.011 & 0.016 & \underline{0.013} & 0.010 & 0.007 & 0.169 & 0.152 & 0.159 & 0.158 & 0.271 & 0.271 & 0.249 & 0.316 \\
\specialrule{\lightrulewidth}{1pt}{1pt}

\multirow{4}{*}{\makecell[l]{Time-LLM \\ {[2024]}}}
& MSE & 0.362 & 0.398 & 0.430 & 0.442 & 0.268 & 0.329 & 0.368 & \textbf{0.372} & \textbf{0.272} & \textbf{0.310} & \textbf{0.352} & \textbf{0.383} & 0.161 & 0.219 & 0.271 & 0.352 & 0.131 & \underline{0.152} & \textbf{0.160} & \textbf{0.192} & \underline{0.362} & \textbf{0.374} & \textbf{0.385} & \textbf{0.430} & 0.147 & \textbf{0.189} & 0.262 & \textbf{0.304} & \underline{0.088} & \underline{0.184} & \underline{0.345} & \underline{0.950} \\
& MAE & 0.392 & 0.418 & 0.427 & 0.457 & \underline{0.328} & 0.375 & 0.409 & \textbf{0.420} & \textbf{0.334} & \textbf{0.358} & 0.384 & 0.411 & 0.253 & 0.293 & 0.329 & 0.379 & \underline{0.224} & 0.242 & \textbf{0.248} & 0.298 & \textbf{0.248} & \textbf{0.247} & \underline{0.271} & \underline{0.288} & 0.201 & \textbf{0.234} & \underline{0.279} & \textbf{0.316} & \underline{0.208} & \underline{0.305} & \underline{0.428} & \underline{0.725} \\
& DTW & 3.780 & 5.540 & 7.820 & 11.950 & 3.140 & 4.810 & 7.850 & 11.310 & \textbf{2.800} & \textbf{4.350} & 6.270 & 9.830 & 2.340 & 3.940 & 5.540 & \underline{8.810} & 2.410 & 3.440 & 4.670 & 7.620 & 3.630 & \textbf{5.260} & 7.180 & 10.980 & 1.960 & 3.370 & 5.740 & \underline{8.560} & \underline{1.960} & 4.610 & 7.750 & 19.770 \\
& TDI & 0.076 & 0.080 & 0.090 & 0.097 & 0.128 & 0.142 & 0.160 & 0.158 & 0.116 & 0.088 & 0.084 & 0.075 & 0.116 & 0.119 & 0.122 & 0.140 & 0.024 & \textbf{0.015} & \textbf{0.012} & \textbf{0.010} & 0.016 & \underline{0.013} & \underline{0.010} & \underline{0.007} & 0.159 & 0.154 & 0.161 & 0.159 & 0.244 & 0.246 & 0.237 & 0.239 \\
\specialrule{\lightrulewidth}{1pt}{1pt}

\multirow{4}{*}{\makecell[l]{GPT4TS \\ {[2023]}}}
& MSE & 0.376 & 0.416 & 0.442 & 0.477 & 0.285 & 0.354 & 0.373 & 0.406 & 0.292 & 0.332 & 0.366 & 0.417 & 0.173 & 0.229 & 0.286 & 0.378 & 0.139 & 0.153 & 0.169 & 0.206 & 0.388 & 0.407 & 0.412 & 0.450 & 0.162 & 0.204 & 0.254 & 0.326 & 0.100 & 0.212 & 0.395 & 1.222 \\
& MAE & 0.397 & 0.418 & 0.433 & \underline{0.456} & 0.342 & 0.389 & 0.407 & 0.441 & 0.346 & 0.372 & 0.394 & 0.421 & 0.262 & 0.301 & 0.341 & 0.401 & 0.238 & 0.251 & 0.266 & 0.297 & 0.282 & 0.290 & 0.294 & 0.312 & 0.212 & 0.248 & 0.286 & 0.337 & 0.221 & 0.328 & 0.455 & 0.838 \\
& DTW & 3.783 & \underline{5.431} & \underline{7.370} & \underline{11.316} & 3.184 & \underline{4.682} & \underline{6.628} & \textbf{10.091} & 2.896 & 4.576 & \textbf{6.154} & 9.866 & 2.384 & 3.832 & 5.555 & 9.085 & 2.270 & 3.380 & 4.670 & 7.680 & 3.730 & 5.560 & \underline{7.180} & 11.020 & \underline{1.917} & \underline{3.303} & \underline{4.949} & 8.661 & 2.056 & 4.393 & 8.714 & 23.410 \\
& TDI & 0.075 & 0.070 & 0.069 & 0.073 & 0.120 & 0.133 & \underline{0.132} & 0.136 & \underline{0.113} & 0.088 & \underline{0.078} & 0.075 & 0.119 & 0.120 & 0.124 & 0.147 & \underline{0.022} & 0.017 & 0.014 & 0.015 & \textbf{0.015} & 0.014 & 0.010 & \underline{0.007} & 0.168 & 0.150 & 0.153 & 0.159 & 0.256 & 0.266 & 0.225 & 0.292 \\
\midrule

\multirow{4}{*}{\makecell[l]{PatchTST \\ {[2023]}}}
& MSE & 0.370 & 0.413 & 0.422 & 0.447 & 0.274 & 0.339 & \textbf{0.329} & 0.379 & 0.290 & 0.332 & 0.366 & 0.416 & 0.165 & 0.220 & 0.274 & 0.362 & \textbf{0.129} & 0.157 & 0.163 & 0.197 & \textbf{0.360} & \underline{0.379} & \underline{0.392} & \underline{0.432} & 0.149 & 0.194 & \underline{0.245} & 0.314 & 0.107 & 0.250 & 0.392 & 1.090 \\
& MAE & 0.399 & 0.421 & 0.436 & 0.466 & 0.336 & 0.379 & \textbf{0.380} & 0.422 & 0.342 & 0.369 & 0.392 & 0.420 & 0.255 & 0.292 & 0.329 & 0.385 & \textbf{0.222} & \textbf{0.240} & 0.259 & \textbf{0.290} & \underline{0.249} & \underline{0.256} & \textbf{0.264} & \textbf{0.286} & \underline{0.198} & 0.241 & 0.282 & 0.334 & 0.234 & 0.362 & 0.465 & 0.777 \\
& DTW & 3.870 & 5.742 & 7.775 & 12.089 & 3.216 & 4.963 & 7.027 & 11.521 & 2.984 & 4.662 & 6.312 & \textbf{9.590} & 2.398 & 3.958 & 5.609 & 8.852 & 2.376 & 3.554 & 4.662 & 7.618 & \textbf{3.610} & 5.332 & 7.318 & 11.031 & 1.994 & 3.464 & 5.370 & 8.850 & 2.093 & 4.801 & 7.102 & 19.701 \\
& TDI & 0.078 & 0.083 & 0.089 & 0.099 & 0.131 & 0.147 & 0.147 & 0.162 & 0.114 & 0.094 & 0.088 & 0.081 & 0.119 & 0.120 & 0.124 & 0.135 & 0.023 & 0.016 & 0.013 & \underline{0.011} & 0.016 & \underline{0.013} & 0.011 & \underline{0.007} & 0.161 & 0.158 & 0.151 & 0.158 & 0.257 & 0.242 & 0.297 & 0.221 \\
\specialrule{\lightrulewidth}{1pt}{1pt}

\multirow{4}{*}{\makecell[l]{TimesNet \\ {[2023]}}}
& MSE & 0.384 & 0.436 & 0.491 & 0.521 & 0.340 & 0.402 & 0.452 & 0.462 & 0.338 & 0.374 & 0.410 & 0.478 & 0.187 & 0.249 & 0.321 & 0.408 & 0.168 & 0.184 & 0.198 & 0.220 & 0.593 & 0.617 & 0.629 & 0.640 & 0.172 & 0.219 & 0.280 & 0.365 & 0.103 & 0.226 & 0.367 & 0.964 \\
& MAE & 0.402 & 0.429 & 0.469 & 0.500 & 0.374 & 0.414 & 0.452 & 0.468 & 0.375 & 0.387 & 0.411 & 0.450 & 0.267 & 0.309 & 0.351 & 0.403 & 0.272 & 0.289 & 0.300 & 0.320 & 0.321 & 0.336 & 0.336 & 0.350 & 0.220 & 0.261 & 0.306 & 0.359 & 0.227 & 0.344 & 0.448 & 0.746 \\
& DTW & 4.397 & 6.043 & 8.150 & 13.762 & 3.651 & 5.201 & 8.018 & 12.413 & 3.367 & 5.183 & 7.289 & 10.560 & 2.450 & 4.250 & 6.150 & 9.550 & 2.608 & 3.824 & 5.391 & 7.842 & 4.020 & 6.060 & 8.090 & 12.810 & 2.290 & 3.600 & 5.620 & 9.650 & 1.996 & \underline{4.340} & \underline{6.640} & 17.420 \\
& TDI & 0.093 & 0.097 & 0.115 & 0.106 & 0.141 & 0.152 & 0.207 & 0.202 & 0.120 & 0.106 & 0.102 & 0.098 & 0.122 & 0.125 & 0.145 & 0.165 & 0.026 & 0.018 & 0.016 & 0.013 & 0.018 & 0.015 & 0.012 & \underline{0.007} & 0.178 & 0.163 & 0.169 & 0.185 & 0.244 & 0.249 & 0.278 & 0.245 \\
\specialrule{\lightrulewidth}{1pt}{1pt}

\multirow{4}{*}{\makecell[l]{DLinear \\ {[2023]}}}
& MSE & 0.375 & 0.405 & 0.439 & 0.472 & 0.289 & 0.383 & 0.448 & 0.605 & 0.299 & 0.335 & 0.369 & 0.425 & 0.167 & 0.224 & 0.281 & 0.397 & 0.140 & 0.153 & 0.169 & 0.203 & 0.410 & 0.423 & 0.436 & 0.466 & 0.176 & 0.220 & 0.265 & 0.333 & 0.090 & 0.186 & 0.348 & 0.955 \\
& MAE & 0.399 & 0.416 & 0.443 & 0.490 & 0.353 & 0.418 & 0.465 & 0.551 & 0.343 & 0.365 & 0.386 & 0.421 & 0.269 & 0.303 & 0.342 & 0.421 & 0.237 & 0.249 & 0.267 & 0.301 & 0.282 & 0.287 & 0.296 & 0.315 & 0.237 & 0.282 & 0.319 & 0.362 & 0.210 & 0.307 & 0.431 & 0.730 \\
& DTW & \underline{3.705} & 5.557 & 7.726 & 12.045 & \underline{3.048} & 4.989 & 6.784 & 11.548 & 3.021 & 4.501 & 6.312 & 10.350 & 2.422 & 3.768 & 5.415 & 9.291 & \textbf{2.233} & \textbf{3.325} & 4.672 & \underline{7.571} & 3.656 & 5.366 & 7.266 & 11.103 & 2.195 & 3.620 & 5.397 & 9.192 & 2.004 & 4.663 & 7.818 & 19.881 \\
& TDI & 0.074 & 0.070 & 0.084 & 0.106 & 0.117 & \underline{0.128} & 0.139 & 0.197 & 0.113 & \textbf{0.088} & 0.082 & 0.087 & 0.120 & 0.118 & \underline{0.117} & 0.138 & 0.023 & 0.017 & 0.014 & 0.014 & 0.016 & 0.013 & 0.011 & 0.008 & 0.186 & 0.164 & 0.167 & 0.176 & 0.250 & 0.249 & 0.240 & 0.241 \\
\midrule

\multirow{4}{*}{\makecell[l]{Stationary \\ {[2022]}}}
& MSE & 0.513 & 0.534 & 0.588 & 0.643 & 0.476 & 0.512 & 0.552 & 0.562 & 0.386 & 0.459 & 0.495 & 0.585 & 0.192 & 0.280 & 0.334 & 0.417 & 0.169 & 0.182 & 0.200 & 0.222 & 0.612 & 0.613 & 0.618 & 0.653 & 0.173 & 0.245 & 0.321 & 0.414 & 0.207 & 0.422 & 1.203 & 2.024 \\
& MAE & 0.491 & 0.504 & 0.535 & 0.616 & 0.458 & 0.493 & 0.551 & 0.560 & 0.398 & 0.444 & 0.464 & 0.516 & 0.274 & 0.339 & 0.361 & 0.413 & 0.273 & 0.286 & 0.304 & 0.321 & 0.338 & 0.340 & 0.328 & 0.355 & 0.223 & 0.285 & 0.338 & 0.410 & 0.340 & 0.478 & 0.821 & 0.984 \\
& DTW & 4.714 & 7.245 & 9.398 & 13.527 & 3.831 & 5.443 & 7.492 & 11.170 & 3.767 & 5.666 & 7.805 & 10.923 & 3.437 & 5.172 & 6.503 & 10.403 & 2.762 & 3.966 & 5.252 & 7.759 & 4.118 & 6.025 & 7.949 & 13.078 & 2.313 & 3.959 & 5.745 & 8.778 & 3.134 & 6.307 & 14.475 & 26.748 \\
& TDI & 0.106 & 0.129 & 0.113 & 0.143 & 0.177 & 0.185 & 0.189 & 0.161 & 0.141 & 0.119 & 0.115 & 0.101 & 0.164 & 0.182 & 0.157 & 0.162 & 0.028 & 0.019 & 0.015 & 0.013 & 0.019 & 0.015 & 0.011 & 0.007 & 0.179 & 0.154 & 0.160 & 0.157 & 0.240 & 0.252 & 0.198 & 0.203 \\
\specialrule{\lightrulewidth}{1pt}{1pt}

\multirow{4}{*}{\makecell[l]{FEDformer \\ {[2022]}}}
& MSE & 0.376 & 0.420 & 0.459 & 0.506 & 0.358 & 0.429 & 0.496 & 0.463 & 0.379 & 0.426 & 0.445 & 0.543 & 0.203 & 0.269 & 0.325 & 0.421 & 0.193 & 0.201 & 0.214 & 0.246 & 0.587 & 0.604 & 0.621 & 0.626 & 0.217 & 0.276 & 0.339 & 0.403 & 0.392 & 0.865 & 0.868 & 1.940 \\
& MAE & 0.419 & 0.448 & 0.465 & 0.507 & 0.397 & 0.439 & 0.487 & 0.474 & 0.419 & 0.441 & 0.459 & 0.490 & 0.287 & 0.328 & 0.366 & 0.415 & 0.308 & 0.315 & 0.329 & 0.355 & 0.366 & 0.373 & 0.383 & 0.382 & 0.296 & 0.336 & 0.380 & 0.428 & 0.475 & 0.736 & 0.733 & 1.099 \\
& DTW & 4.122 & 6.157 & 7.924 & 12.000 & 3.904 & 5.901 & 7.354 & 12.581 & 4.342 & 5.413 & 7.530 & 11.420 & 3.299 & 4.954 & 6.610 & 11.107 & 3.058 & 4.379 & 5.932 & 8.780 & 4.422 & 6.510 & 10.367 & 12.309 & 2.330 & 5.814 & 6.965 & 10.293 & 4.715 & 10.264 & 13.639 & 31.102 \\
& TDI & 0.088 & 0.088 & \underline{0.069} & 0.079 & 0.155 & 0.185 & 0.163 & 0.181 & 0.148 & 0.104 & 0.093 & 0.078 & 0.133 & 0.133 & 0.138 & 0.178 & 0.030 & 0.021 & 0.017 & 0.013 & 0.021 & 0.015 & 0.011 & 0.008 & \underline{0.159} & \textbf{0.132} & \textbf{0.145} & \textbf{0.146} & \underline{0.168} & \textbf{0.160} & \textbf{0.173} & \textbf{0.167} \\
\bottomrule
\end{tabular}%
}
\end{table*}

\begin{table*}
\section{Few-Shot Forecasting}
\label{app:full_few-shot_results}

\centering
\scriptsize
\setlength{\tabcolsep}{2.5pt}
\renewcommand{\arraystretch}{1.0}
\caption{Full few-shot learning results on 10\% training data. We use the same protocol in Table~\ref{tab:main_results_avg}.}
\label{tab:main_results_fewshot_top_conference}

\fontsize{11pt}{12pt}\selectfont

\resizebox{\textwidth}{!}{%
\begin{tabular}{ll cccc cccc cccc cccc cccc cccc cccc cccc}
\toprule
\multirow{2}{*}{\textbf{Models}} & \multirow{2}{*}{\textbf{Metrics}}
& \multicolumn{4}{c}{\textbf{ETTh1}}
& \multicolumn{4}{c}{\textbf{ETTh2}}
& \multicolumn{4}{c}{\textbf{ETTm1}}
& \multicolumn{4}{c}{\textbf{ETTm2}}
& \multicolumn{4}{c}{\textbf{Electricity}}
& \multicolumn{4}{c}{\textbf{Traffic}}
& \multicolumn{4}{c}{\textbf{Weather}}
& \multicolumn{4}{c}{\textbf{Exchange}} \\
\cmidrule(lr){3-6} \cmidrule(lr){7-10} \cmidrule(lr){11-14} \cmidrule(lr){15-18}
\cmidrule(lr){19-22} \cmidrule(lr){23-26} \cmidrule(lr){27-30} \cmidrule(lr){31-34}
& & \textbf{96} & \textbf{192} & \textbf{336} & \textbf{720} & \textbf{96} & \textbf{192} & \textbf{336} & \textbf{720} & \textbf{96} & \textbf{192} & \textbf{336} & \textbf{720} & \textbf{96} & \textbf{192} & \textbf{336} & \textbf{720} & \textbf{96} & \textbf{192} & \textbf{336} & \textbf{720} & \textbf{96} & \textbf{192} & \textbf{336} & \textbf{720} & \textbf{96} & \textbf{192} & \textbf{336} & \textbf{720} & \textbf{96} & \textbf{192} & \textbf{336} & \textbf{720} \\
\midrule

\multirow{4}{*}{\makecell[l]{STaT \\ \textbf{[Ours]}}}
& MSE & \textbf{0.369} & \textbf{0.400} & \textbf{0.418} & \textbf{0.451} & \underline{0.277} & \textbf{0.337} & \textbf{0.366} & \textbf{0.411} & \textbf{0.291} & \textbf{0.330} & \textbf{0.362} & \underline{0.428} & \textbf{0.165} & \textbf{0.220} & \underline{0.277} & \textbf{0.354} & \textbf{0.144} & \textbf{0.162} & \underline{0.178} & \textbf{0.222} & \underline{0.409} & 0.417 & 0.439 & 0.476 & \textbf{0.152} & \textbf{0.200} & \textbf{0.246} & \underline{0.319} & \textbf{0.104} & \textbf{0.200} & \underline{0.353} & \textbf{0.912} \\
& MAE & \textbf{0.400} & \textbf{0.418} & \textbf{0.432} & \textbf{0.462} & \underline{0.340} & \underline{0.379} & \textbf{0.402} & \textbf{0.445} & \textbf{0.342} & \textbf{0.365} & \underline{0.390} & \underline{0.425} & \textbf{0.255} & \textbf{0.292} & \underline{0.332} & \textbf{0.382} & \textbf{0.246} & \textbf{0.258} & \underline{0.273} & \textbf{0.315} & 0.298 & 0.306 & 0.306 & 0.332 & \textbf{0.201} & \textbf{0.246} & \textbf{0.282} & \underline{0.335} & \textbf{0.230} & \textbf{0.319} & \underline{0.433} & \textbf{0.725} \\
& DTW & \underline{3.784} & \underline{5.509} & \textbf{7.404} & \textbf{10.934} & \underline{3.200} & \underline{5.073} & \textbf{6.471} & \underline{10.294} & \textbf{2.950} & \textbf{4.360} & \textbf{5.860} & \textbf{9.283} & \textbf{2.258} & \textbf{3.697} & \textbf{5.333} & \textbf{8.842} & \underline{2.322} & 3.488 & 4.848 & \textbf{7.020} & 3.885 & 5.601 & \textbf{7.214} & \underline{10.778} & \textbf{1.930} & \underline{3.328} & \textbf{4.980} & \textbf{8.669} & \underline{2.283} & \underline{4.410} & \underline{7.408} & \underline{14.095} \\
& TDI & \underline{0.071} & \textbf{0.064} & \textbf{0.063} & \underline{0.058} & \textbf{0.114} & \textbf{0.127} & \textbf{0.125} & \underline{0.145} & \textbf{0.109} & \textbf{0.086} & \textbf{0.074} & \textbf{0.071} & \textbf{0.116} & \textbf{0.113} & \textbf{0.113} & \textbf{0.133} & \underline{0.023} & \underline{0.017} & \textbf{0.013} & \textbf{0.011} & \underline{0.018} & \underline{0.014} & \underline{0.011} & \underline{0.008} & \textbf{0.160} & \underline{0.150} & \underline{0.151} & \underline{0.157} & \underline{0.182} & \underline{0.229} & \underline{0.209} & \underline{0.217} \\
\specialrule{\lightrulewidth}{1pt}{1pt}

\multirow{4}{*}{\makecell[l]{TimeCMA \\ {[2025]}}}
& MSE & 0.422 & 0.483 & 0.540 & 0.545 & 0.345 & 0.442 & 0.476 & 0.465 & 0.356 & 0.408 & 0.450 & 0.525 & 0.196 & 0.263 & 0.325 & 0.424 & 0.169 & 0.186 & 0.205 & 0.269 & 0.499 & 0.534 & 0.566 & 0.638 & 0.172 & 0.221 & 0.272 & 0.366 & 0.120 & 0.227 & 0.429 & 1.222 \\
& MAE & 0.430 & 0.463 & 0.485 & 0.502 & 0.376 & 0.431 & 0.460 & 0.469 & 0.388 & 0.414 & 0.436 & 0.473 & 0.274 & 0.317 & 0.355 & 0.410 & 0.272 & 0.285 & 0.304 & 0.359 & 0.348 & 0.371 & 0.388 & 0.426 & 0.219 & 0.262 & 0.298 & 0.358 & 0.246 & 0.345 & 0.479 & 0.836 \\
& DTW & 3.883 & 5.833 & 8.190 & 11.907 & 3.303 & 5.159 & 7.324 & 10.769 & 3.213 & 4.815 & 6.619 & 10.750 & 2.479 & 3.973 & 5.773 & 9.820 & 2.525 & \underline{3.703} & \underline{5.160} & \underline{7.282} & 4.025 & 6.011 & 7.717 & \textbf{8.875} & 2.126 & 3.690 & 5.652 & 10.038 & 2.297 & 4.527 & 7.714 & 16.884 \\
& TDI & \textbf{0.070} & \textbf{0.064} & \underline{0.066} & \textbf{0.057} & \underline{0.115} & \underline{0.133} & \underline{0.139} & \textbf{0.136} & \underline{0.125} & 0.097 & \underline{0.080} & \underline{0.073} & \underline{0.128} & \underline{0.123} & \underline{0.129} & \underline{0.134} & 0.025 & \underline{0.017} & \underline{0.014} & 0.013 & 0.021 & 0.018 & 0.014 & 0.012 & 0.178 & 0.160 & 0.166 & 0.167 & 0.235 & 0.245 & 0.264 & 0.331 \\
\specialrule{\lightrulewidth}{1pt}{1pt}

\multirow{4}{*}{\makecell[l]{Time-VLM \\ {[2025]}}}
& MSE & \underline{0.391} & \underline{0.420} & \underline{0.439} & \underline{0.476} & 0.284 & \underline{0.349} & \underline{0.370} & 0.441 & \underline{0.310} & \underline{0.340} & \underline{0.369} & \textbf{0.423} & \underline{0.169} & \underline{0.222} & 0.278 & \underline{0.381} & 0.160 & 0.174 & 0.190 & 0.229 & 0.465 & 0.468 & 0.483 & 0.520 & 0.174 & 0.217 & 0.263 & 0.326 & 0.111 & 0.208 & 0.399 & 1.377 \\
& MAE & \underline{0.404} & \underline{0.431} & \underline{0.448} & \underline{0.484} & 0.347 & 0.398 & \underline{0.412} & 0.466 & \underline{0.354} & \underline{0.370} & \textbf{0.387} & \textbf{0.417} & \underline{0.260} & \underline{0.296} & 0.335 & 0.401 & 0.269 & 0.279 & 0.294 & 0.323 & 0.349 & 0.350 & 0.356 & 0.373 & 0.228 & 0.262 & 0.296 & 0.340 & 0.236 & 0.326 & 0.462 & 0.876 \\
& DTW & 4.002 & 5.841 & 7.973 & 11.821 & 3.405 & 5.375 & 8.179 & 13.070 & 3.115 & 4.645 & 6.478 & 10.339 & 2.558 & 3.991 & 5.721 & 9.232 & 2.607 & 3.815 & 5.316 & 8.497 & 4.224 & 6.083 & \underline{8.544} & \underline{12.591} & 2.069 & 3.475 & 5.283 & 9.018 & 2.323 & 4.735 & 8.645 & 17.797 \\
& TDI & 0.073 & \underline{0.067} & \underline{0.066} & 0.069 & 0.135 & 0.147 & 0.182 & 0.211 & 0.117 & \underline{0.089} & 0.080 & \underline{0.073} & 0.124 & 0.121 & 0.127 & 0.137 & 0.024 & \underline{0.017} & \textbf{0.013} & 0.012 & \underline{0.018} & \underline{0.014} & \underline{0.011} & \underline{0.008} & 0.174 & 0.154 & 0.159 & 0.161 & 0.224 & 0.329 & 0.289 & 0.320 \\
\specialrule{\lightrulewidth}{1pt}{1pt}

\multirow{4}{*}{\makecell[l]{Time-LLM \\ {[2024]}}}
& MSE & 0.448 & 0.484 & 0.589 & 0.700 & \textbf{0.275} & 0.374 & 0.406 & \underline{0.427} & 0.346 & 0.373 & 0.413 & 0.485 & 0.177 & 0.241 & \textbf{0.274} & 0.417 & \textbf{0.139} & \textbf{0.151} & \textbf{0.169} & 0.240 & 0.418 & \textbf{0.414} & \textbf{0.421} & \textbf{0.462} & \underline{0.161} & \underline{0.204} & 0.261 & \textbf{0.309} & 0.121 & 0.234 & 0.377 & 0.954 \\
& MAE & 0.460 & 0.483 & 0.540 & 0.604 & \textbf{0.326} & \textbf{0.373} & \underline{0.429} & \underline{0.449} & 0.388 & 0.416 & 0.426 & 0.476 & 0.261 & 0.314 & \textbf{0.327} & \underline{0.390} & \textbf{0.241} & \textbf{0.248} & \textbf{0.270} & 0.322 & \underline{0.291} & \textbf{0.296} & 0.311 & 0.327 & \underline{0.210} & \underline{0.248} & 0.302 & \textbf{0.332} & 0.291 & \underline{0.324} & 0.454 & 0.857 \\
& DTW & 5.534 & 8.613 & 8.248 & 14.748 & 3.681 & 5.517 & 6.914 & 11.476 & 3.558 & 7.798 & 10.481 & 10.215 & 2.577 & 4.705 & 5.949 & 9.711 & \textbf{2.350} & \underline{3.450} & \underline{4.750} & 7.850 & \textbf{3.850} & \textbf{5.550} & \textbf{7.350} & \textbf{10.950} & \underline{1.950} & \textbf{3.320} & 5.260 & 8.850 & 2.591 & 4.453 & 8.374 & 15.588 \\
& TDI & 0.106 & 0.170 & 0.112 & 0.076 & 0.141 & \underline{0.133} & \underline{0.131} & 0.191 & 0.118 & 0.171 & 0.142 & 0.075 & \underline{0.119} & 0.141 & 0.137 & 0.151 & 0.023 & \textbf{0.016} & \textbf{0.013} & 0.012 & \underline{0.017} & \textbf{0.013} & \underline{0.011} & \textbf{0.007} & 0.168 & 0.151 & 0.155 & 0.158 & 0.244 & 0.238 & 0.291 & 0.248 \\
\specialrule{\lightrulewidth}{1pt}{1pt}

\multirow{4}{*}{\makecell[l]{GPT4TS \\ {[2023]}}}
& MSE & 0.458 & 0.570 & 0.608 & 0.725 & 0.331 & 0.402 & 0.406 & 0.449 & 0.390 & 0.429 & 0.469 & 0.569 & 0.188 & 0.251 & 0.307 & 0.426 & \textbf{0.139} & \underline{0.156} & \underline{0.175} & 0.233 & 0.414 & 0.426 & 0.434 & 0.487 & 0.163 & 0.210 & \underline{0.256} & 0.321 & 0.110 & 0.214 & 0.407 & 1.094 \\
& MAE & 0.456 & 0.516 & 0.535 & 0.591 & 0.374 & 0.411 & 0.433 & 0.464 & 0.404 & 0.423 & 0.439 & 0.498 & 0.269 & 0.309 & 0.346 & 0.417 & \textbf{0.237} & \underline{0.252} & \textbf{0.270} & \underline{0.317} & 0.297 & \underline{0.301} & \textbf{0.303} & 0.337 & 0.215 & 0.254 & \underline{0.292} & 0.339 & \underline{0.233} & 0.330 & 0.465 & 0.790 \\
& DTW & 3.860 & 5.707 & \underline{7.728} & 11.695 & 3.292 & 5.325 & 7.174 & 11.423 & \underline{3.016} & 4.542 & \underline{6.309} & \underline{10.123} & 2.434 & 3.864 & \underline{5.600} & \underline{9.003} & \textbf{2.247} & \textbf{3.331} & \textbf{4.671} & 7.591 & \underline{3.820} & 5.650 & 7.550 & 11.850 & 2.150 & 3.478 & 5.273 & 9.030 & 2.370 & 4.723 & 8.770 & 22.722 \\
& TDI & 0.073 & \underline{0.067} & 0.069 & 0.069 & 0.119 & 0.150 & 0.147 & 0.163 & 0.115 & \underline{0.090} & \underline{0.078} & \underline{0.073} & 0.120 & \underline{0.118} & \underline{0.125} & 0.149 & \textbf{0.022} & \textbf{0.016} & \textbf{0.013} & \textbf{0.011} & \textbf{0.017} & \textbf{0.013} & \textbf{0.011} & \underline{0.007} & 0.170 & 0.152 & 0.156 & 0.159 & 0.227 & 0.275 & 0.238 & 0.264 \\
\midrule

\multirow{4}{*}{\makecell[l]{PatchTST \\ {[2023]}}}
& MSE & 0.516 & 0.598 & 0.657 & 0.762 & 0.353 & 0.403 & 0.426 & 0.477 & 0.410 & 0.437 & 0.476 & 0.681 & 0.191 & 0.252 & 0.306 & 0.433 & \underline{0.140} & 0.160 & 0.180 & 0.241 & \textbf{0.403} & \underline{0.415} & \underline{0.426} & \underline{0.474} & 0.165 & 0.210 & 0.259 & 0.332 & \underline{0.108} & \underline{0.207} & 0.358 & 1.409 \\
& MAE & 0.485 & 0.524 & 0.550 & 0.610 & 0.389 & 0.414 & 0.441 & 0.480 & 0.419 & 0.434 & 0.454 & 0.556 & 0.274 & 0.317 & 0.353 & 0.427 & \underline{0.238} & \underline{0.255} & \underline{0.276} & 0.323 & \textbf{0.289} & \textbf{0.296} & \underline{0.304} & \underline{0.331} & 0.215 & 0.257 & 0.297 & 0.346 & 0.236 & 0.330 & \textbf{0.412} & 0.895 \\
& DTW & \textbf{3.782} & \textbf{5.453} & \textbf{7.994} & \underline{11.656} & \textbf{3.069} & \textbf{4.774} & \underline{6.541} & \underline{10.513} & \textbf{3.121} & \textbf{4.687} & \textbf{6.606} & \textbf{10.146} & \underline{2.408} & \underline{3.779} & \underline{5.655} & 10.700 & 2.428 & 3.591 & 5.504 & 7.936 & \textbf{3.811} & \underline{5.561} & \underline{7.318} & 11.150 & 1.989 & 3.370 & \underline{5.250} & 9.000 & \textbf{2.152} & \textbf{4.088} & \textbf{6.992} & 20.198 \\
& TDI & 0.076 & 0.070 & 0.082 & 0.065 & 0.128 & 0.136 & 0.137 & 0.160 & 0.116 & 0.093 & 0.083 & 0.077 & 0.123 & 0.120 & 0.132 & 0.157 & \underline{0.023} & \textbf{0.016} & \underline{0.014} & \textbf{0.011} & \textbf{0.017} & \textbf{0.013} & \underline{0.011} & \textbf{0.007} & 0.169 & 0.153 & 0.152 & 0.161 & 0.246 & 0.255 & 0.264 & 0.338 \\
\specialrule{\lightrulewidth}{1pt}{1pt}

\multirow{4}{*}{\makecell[l]{TimesNet \\ {[2023]}}}
& MSE & 0.861 & 0.797 & 0.941 & 0.877 & 0.378 & 0.490 & 0.537 & 0.510 & 0.583 & 0.630 & 0.725 & 0.769 & 0.212 & 0.270 & 0.323 & 0.474 & 0.299 & 0.305 & 0.319 & 0.369 & 0.719 & 0.748 & 0.853 & 1.485 & 0.184 & 0.245 & 0.305 & 0.381 & 0.285 & 0.386 & 1.232 & 1.764 \\
& MAE & 0.628 & 0.593 & 0.648 & 0.641 & 0.409 & 0.467 & 0.494 & 0.491 & 0.501 & 0.528 & 0.568 & 0.549 & 0.285 & 0.323 & 0.353 & 0.449 & 0.373 & 0.379 & 0.391 & 0.426 & 0.416 & 0.428 & 0.471 & 0.825 & 0.230 & 0.283 & 0.321 & 0.371 & 0.394 & 0.466 & 0.853 & 1.026 \\
& DTW & 4.462 & 6.164 & 8.721 & 12.534 & 3.708 & 5.556 & 9.687 & 13.487 & 3.240 & 4.937 & 7.374 & 10.796 & 2.781 & 4.208 & 7.210 & 14.020 & 3.087 & 4.411 & 6.033 & 9.303 & 4.406 & 6.349 & 9.898 & 14.614 & 2.052 & 3.738 & 5.852 & 8.960 & 3.812 & 6.366 & 15.111 & 24.446 \\
& TDI & 0.095 & 0.085 & 0.091 & 0.080 & 0.153 & 0.159 & 0.206 & 0.223 & \underline{0.113} & 0.095 & 0.087 & \underline{0.073} & 0.121 & 0.122 & 0.202 & 0.256 & 0.032 & 0.023 & 0.019 & 0.016 & 0.019 & \textbf{0.013} & 0.012 & 0.010 & \underline{0.161} & 0.162 & 0.160 & 0.161 & 0.183 & 0.234 & 0.223 & 0.305 \\
\specialrule{\lightrulewidth}{1pt}{1pt}

\multirow{4}{*}{\makecell[l]{DLinear \\ {[2023]}}}
& MSE & 0.492 & 0.565 & 0.721 & 0.986 & 0.357 & 0.569 & 0.671 & 0.824 & 0.352 & 0.382 & 0.419 & 0.490 & 0.213 & 0.278 & 0.338 & 0.436 & 0.150 & 0.164 & 0.181 & \underline{0.223} & 0.419 & 0.434 & 0.449 & 0.484 & 0.171 & 0.215 & 0.258 & 0.320 & 0.157 & 0.219 & \textbf{0.349} & \underline{0.913} \\
& MAE & 0.495 & 0.538 & 0.622 & 0.743 & 0.411 & 0.519 & 0.572 & 0.648 & 0.392 & 0.412 & 0.434 & 0.477 & 0.303 & 0.345 & 0.385 & 0.440 & 0.253 & 0.264 & 0.282 & 0.321 & 0.298 & 0.305 & 0.313 & 0.336 & 0.224 & 0.263 & 0.299 & 0.346 & 0.296 & 0.358 & 0.463 & \underline{0.763} \\
& DTW & 3.820 & 5.648 & 7.806 & 12.087 & 3.216 & 5.436 & 6.736 & 11.450 & 3.040 & \underline{4.539} & 6.329 & 10.133 & 2.417 & 3.855 & \underline{5.374} & 9.600 & 2.367 & 3.504 & 4.880 & 7.852 & 3.895 & 5.711 & 7.725 & 11.773 & 2.167 & 3.578 & 5.383 & 9.145 & 2.765 & 4.413 & 7.674 & \textbf{11.749} \\
& TDI & 0.072 & 0.070 & 0.084 & 0.105 & 0.131 & 0.154 & 0.140 & 0.201 & \underline{0.113} & \underline{0.089} & 0.080 & 0.079 & 0.120 & 0.119 & 0.127 & 0.161 & 0.024 & 0.018 & 0.015 & 0.015 & \textbf{0.017} & \textbf{0.013} & \textbf{0.010} & \textbf{0.007} & 0.182 & 0.163 & 0.166 & 0.174 & 0.222 & 0.256 & 0.218 & 0.299 \\
\midrule

\multirow{4}{*}{\makecell[l]{Stationary \\ {[2022]}}}
& MSE & 0.918 & 0.915 & 0.939 & 0.887 & 0.389 & 0.473 & 0.507 & 0.477 & 0.761 & 0.781 & 0.803 & 0.844 & 0.229 & 0.291 & 0.348 & 0.461 & 0.420 & 0.411 & 0.434 & 0.510 & 1.412 & 1.419 & 1.443 & 1.539 & 0.192 & 0.269 & 0.370 & 0.441 & 0.191 & 0.366 & 0.561 & 1.650 \\
& MAE & 0.639 & 0.629 & 0.644 & 0.645 & 0.411 & 0.455 & 0.480 & 0.472 & 0.568 & 0.574 & 0.587 & 0.581 & 0.308 & 0.343 & 0.376 & 0.438 & 0.466 & 0.459 & 0.473 & 0.521 & 0.802 & 0.806 & 0.815 & 0.837 & 0.234 & 0.295 & 0.357 & 0.405 & 0.330 & 0.457 & 0.563 & 0.946 \\
& DTW & 4.699 & 7.119 & 9.513 & 14.121 & 3.954 & 5.697 & 8.096 & 11.266 & 3.576 & 5.312 & 7.637 & 12.760 & 3.154 & 4.480 & 6.892 & 13.328 & 3.050 & 4.565 & 5.931 & 8.526 & 4.758 & 6.386 & 8.603 & 12.883 & 2.308 & 3.659 & 5.847 & \underline{8.800} & 2.962 & 5.737 & 9.647 & 25.009 \\
& TDI & 0.112 & 0.105 & 0.132 & 0.122 & 0.176 & 0.185 & 0.222 & 0.160 & 0.142 & 0.108 & 0.100 & 0.116 & 0.172 & 0.163 & 0.183 & 0.249 & 0.031 & 0.023 & 0.019 & 0.016 & 0.022 & 0.015 & \textbf{0.010} & \textbf{0.007} & 0.186 & 0.163 & 0.163 & 0.162 & 0.226 & 0.245 & 0.254 & \textbf{0.209} \\
\specialrule{\lightrulewidth}{1pt}{1pt}

\multirow{4}{*}{\makecell[l]{FEDformer \\ {[2022]}}}
& MSE & 0.512 & 0.624 & 0.691 & 0.728 & 0.382 & 0.478 & 0.504 & 0.499 & 0.578 & 0.617 & 0.998 & 0.693 & 0.291 & 0.307 & 0.543 & 0.712 & 0.231 & 0.261 & 0.360 & 0.530 & 0.639 & 0.637 & 0.655 & 0.722 & 0.188 & 0.250 & 0.312 & 0.387 & 0.544 & 0.627 & 0.927 & 1.640 \\
& MAE & 0.499 & 0.555 & 0.574 & 0.614 & 0.416 & 0.474 & 0.501 & 0.509 & 0.518 & 0.546 & 0.775 & 0.579 & 0.399 & 0.379 & 0.559 & 0.614 & 0.323 & 0.356 & 0.445 & 0.585 & 0.400 & 0.416 & 0.427 & 0.456 & 0.253 & 0.304 & 0.346 & 0.393 & 0.556 & 0.608 & 0.752 & 0.980 \\
& DTW & 4.621 & 6.794 & 8.334 & 12.276 & 4.082 & 5.784 & 7.582 & 11.845 & 4.106 & 5.716 & 7.570 & 11.823 & 3.316 & 4.807 & 6.807 & 10.880 & 3.397 & 4.767 & 6.167 & 9.322 & 5.666 & 8.066 & 9.659 & 13.944 & 3.317 & 5.259 & 7.139 & 10.421 & 5.392 & 8.213 & 13.840 & 27.015 \\
& TDI & 0.103 & 0.095 & 0.077 & 0.088 & 0.172 & 0.186 & 0.181 & 0.171 & 0.145 & 0.107 & 0.089 & 0.080 & 0.134 & 0.129 & 0.146 & 0.164 & 0.035 & 0.025 & 0.019 & 0.016 & 0.026 & \underline{0.018} & \underline{0.011} & \underline{0.008} & \textbf{0.169} & \textbf{0.144} & \textbf{0.140} & \textbf{0.144} & \textbf{0.175} & \textbf{0.217} & \textbf{0.188} & 0.233 \\
\bottomrule
\end{tabular}}
\end{table*}

\onecolumn 
\twocolumn

\bibliographystyle{ACM-Reference-Format}
\bibliography{main}
\end{document}